\documentclass[sigconf,natbib=true]{acmart}
\usepackage{adjustbox}
\usepackage{multirow}
\usepackage{makecell}
\AtBeginDocument{%
  }

\setcopyright{acmlicensed}
\copyrightyear{2018}
\acmYear{2018}
\acmDOI{XXXXXXX.XXXXXXX}
\acmConference[Conference acronym 'XX]{Make sure to enter the correct
  conference title from your rights confirmation email}{June 03--05,
  2018}{Woodstock, NY}
\acmISBN{978-1-4503-XXXX-X/2018/06}

\begin{document}

\title{Dual Modality-Aware Gated Prompt Tuning for Few-Shot Multimodal Sarcasm Detection}

\author{Soumyadeep Jana}
\affiliation{%
  \institution{IIT Guwahati}
  \city{Guwahati}
  \state{Assam}
  \country{India}
}
\email{sjana@iitg.ac.in}

\author{Abhrajyoti Kundu}
\affiliation{%
  \institution{IIT Guwahati}
  \city{Guwahati}
  \state{Assam}
  \country{India}
}
\email{abhrajyoti00@gmail.com}

\author{Sanasam Ranbir Singh}
\affiliation{%
  \institution{IIT Guwahati}
  \city{Guwahati}
  \state{Assam}
  \country{India}
}
\email{ranbir@iitg.ac.in}


\begin{abstract}

  The widespread use of multimodal content on social media has heightened the need for effective sarcasm detection to improve opinion mining. However, existing models rely heavily on large annotated datasets, making them less suitable for real-world scenarios where labeled data is scarce. This motivates the need to explore the problem in a few-shot setting. To this end, we introduce DMDP (\textbf{D}eep \textbf{M}odality-\textbf{D}isentangled \textbf{P}rompt Tuning), a novel framework for few-shot multimodal sarcasm detection. Unlike prior methods that use shallow, unified prompts across modalities, DMDP employs gated, modality-specific deep prompts for the text and visual encoders. These prompts are injected across multiple layers to enable hierarchical feature learning and better capture diverse sarcasm types. To enhance intra-modal learning, we incorporate a prompt-sharing mechanism across layers, allowing the model to aggregate both low-level and high-level semantic cues. Additionally, a cross-modal prompt alignment module enables nuanced interactions between image and text representations, improving the model’s ability to detect subtle sarcastic intent. Experiments on two public datasets demonstrate DMDP’s superior performance in both few-shot and extremely low-resource settings. Further cross-dataset evaluations show that DMDP generalizes well across domains, consistently outperforming baseline methods.
\end{abstract}





\maketitle

\section{Introduction}
The proliferation of social media has given people the power to express themselves in creative ways. Often to express discontentment, they resort to irony and sarcasm. With these platforms enabling multimedia content, the use of multimodal (image-text) sarcasm has seen tremendous growth. The detection of multimodal sarcasm is crucial to interpreting users' opinions and sentiments accurately \cite{Tindale1987TheUO, Averbeck2013ComparisonsOI, maynard-greenwood-2014-cares, badlani-etal-2019-ensemble, ghosh-etal-2021-laughing}. This is attributed to the fact that sarcasm tends to flip the polarity of utterances.

Existing works on multimodal image-text sarcasm \cite{cai-etal-2019-multi,xu-etal-2020-reasoning,Liang2021MultiModalSD,liu-etal-2022-towards-multi-modal,tian-etal-2023-dynamic, qin-etal-2023-mmsd2, Tang2024LeveragingGL} require vast amounts of annotated data to perform well due to their overtly complex structures. However, obtaining such datasets is challenging, as the annotation process is resource-intensive and time-consuming. Recent works \cite{Yu2022UnifiedMP, Yu2022FewShotMS, Yang2022FewshotMS, cao-etal-2022-prompting} have explored prompt-learning techniques for tasks like few-shot multimodal sentiment analysis and hate speech detection, where the model backbone is frozen and only the prompts are optimized during training. Prompt-learning is particularly effective in few-shot settings because it leverages the knowledge already encoded in pre-trained models, requiring only minimal task-specific adaptation. Following a similar direction, work by \cite{Jana2024ContinuousAM} presented the first dedicated study on few-shot multimodal image-text sarcasm detection and proposed their model \texttt{CAMP}. In contrast to earlier prompt-learning approaches, \texttt{CAMP} introduced an attentive and dynamic prompt tuning strategy using a BERT \cite{Devlin2019BERTPO} backbone. Their method generated prompts by attending to both image and text tokens, based on the intuition that sarcasm often emerges from the incongruity between visual and textual content. This dynamic mechanism enabled \texttt{CAMP} to achieve notable improvements, significantly outperforming existing baselines.

\begin{figure}[t]
    \centering
\begin{tabular}{p{0.45\linewidth}p{0.45\linewidth}}
 \includegraphics[width=\linewidth, height=\linewidth]{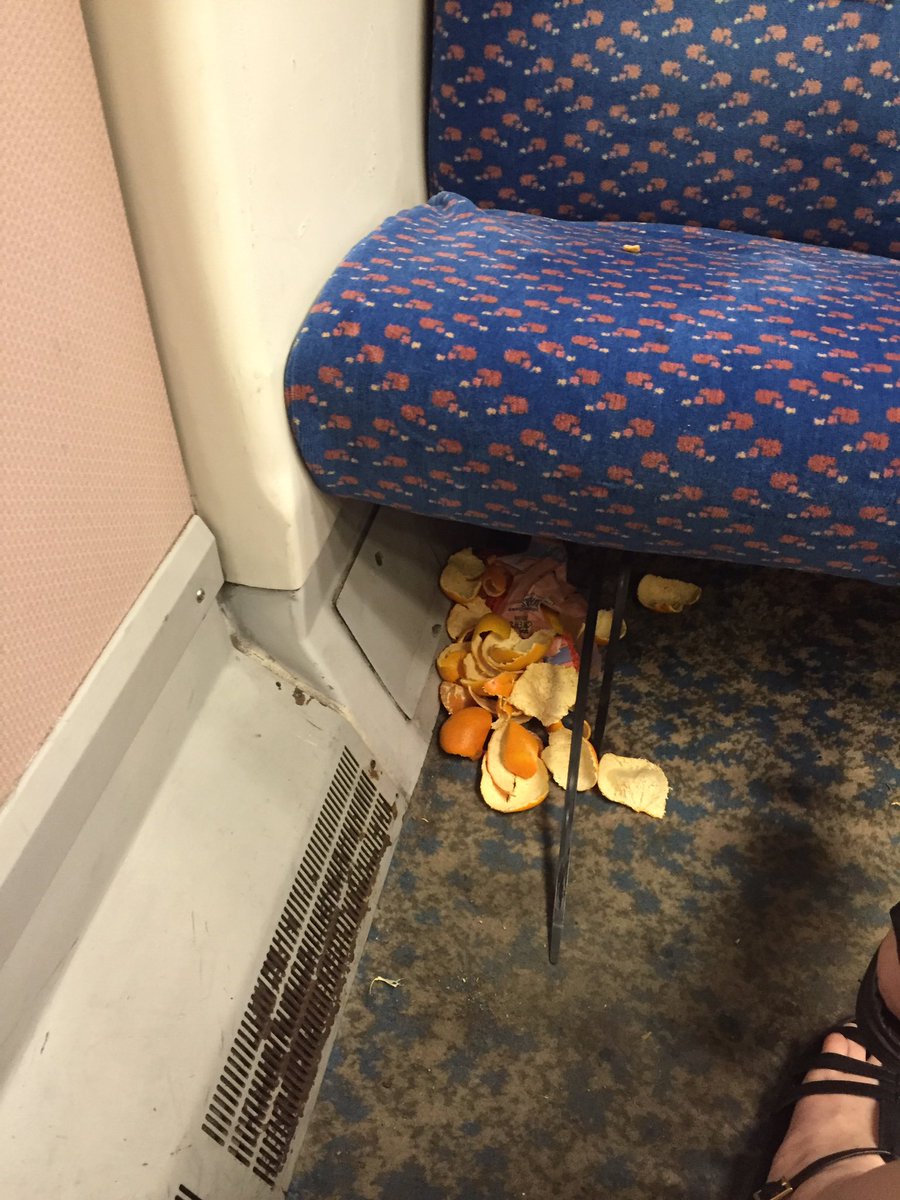} & \includegraphics[width=\linewidth, height=\linewidth]{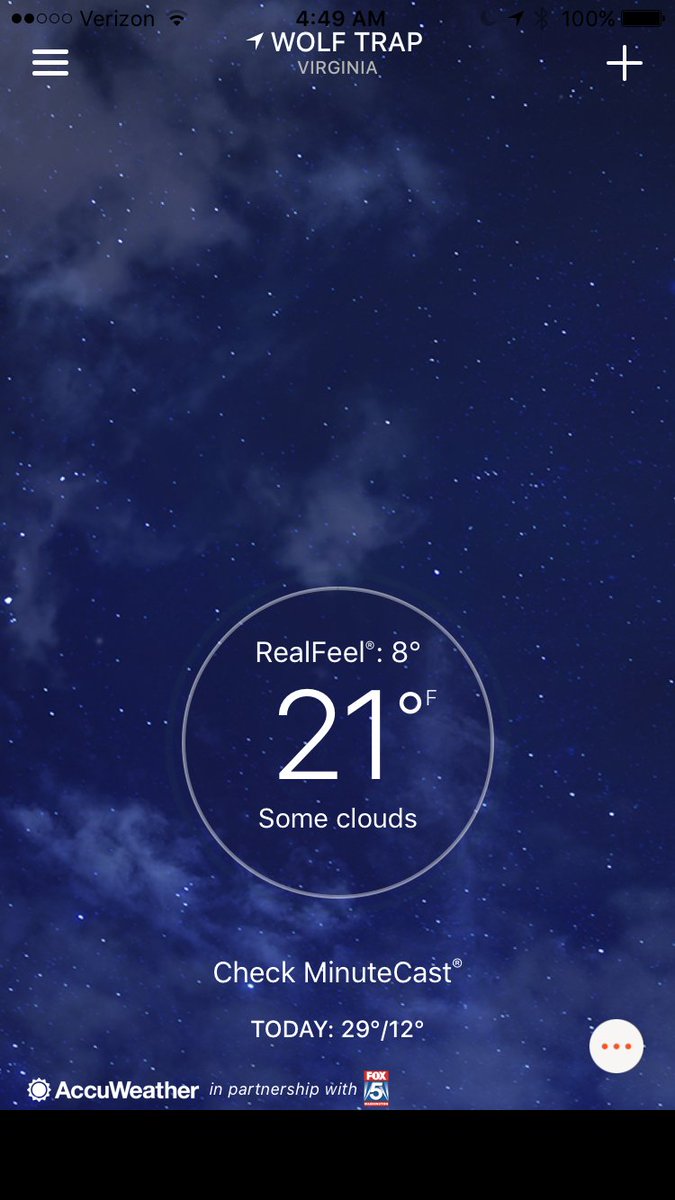} \\
  \footnotesize{(a) lovely, clean, pleasant <user> train home}&  \footnotesize{(b) nice warm running weather this morning} \\
 &\\
 \includegraphics[width=\linewidth, height=\linewidth]{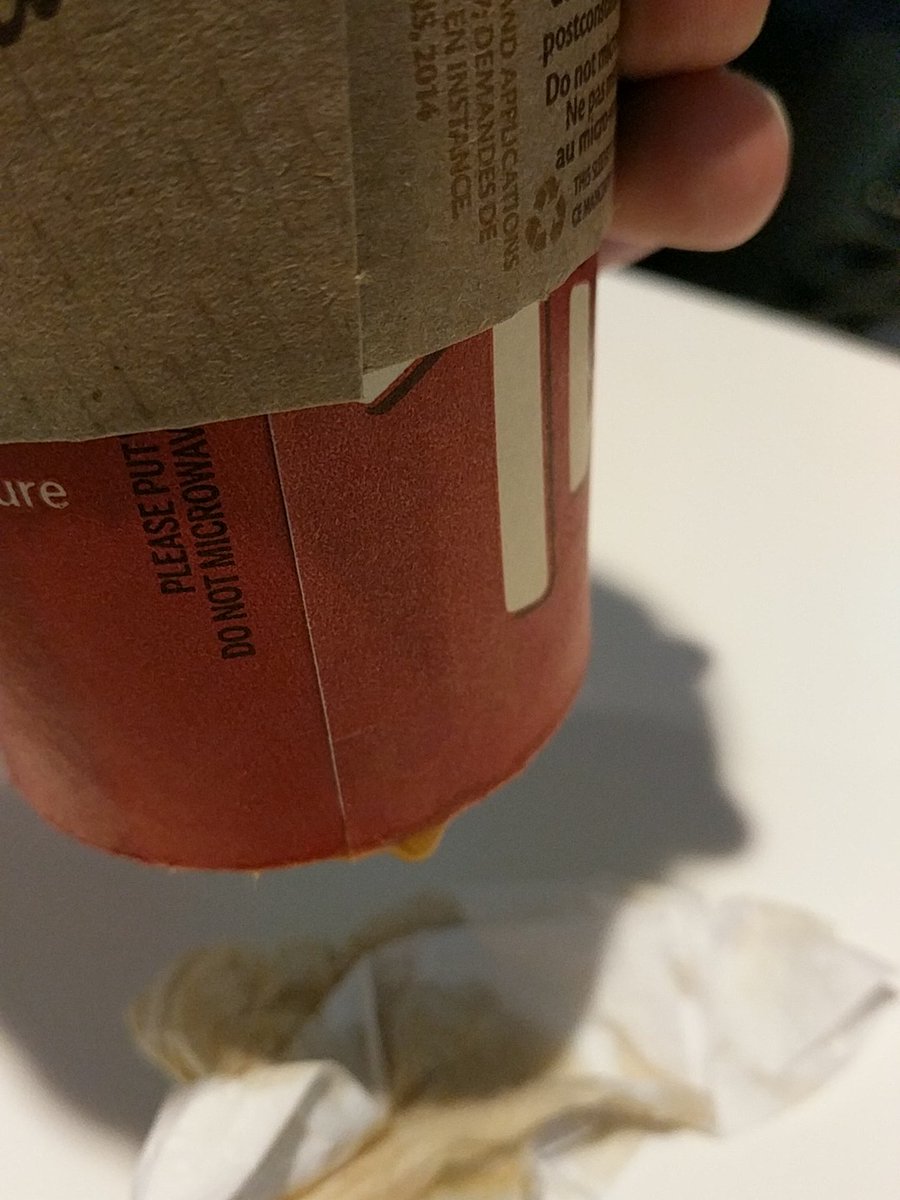}& \includegraphics[width=\linewidth, height=\linewidth]{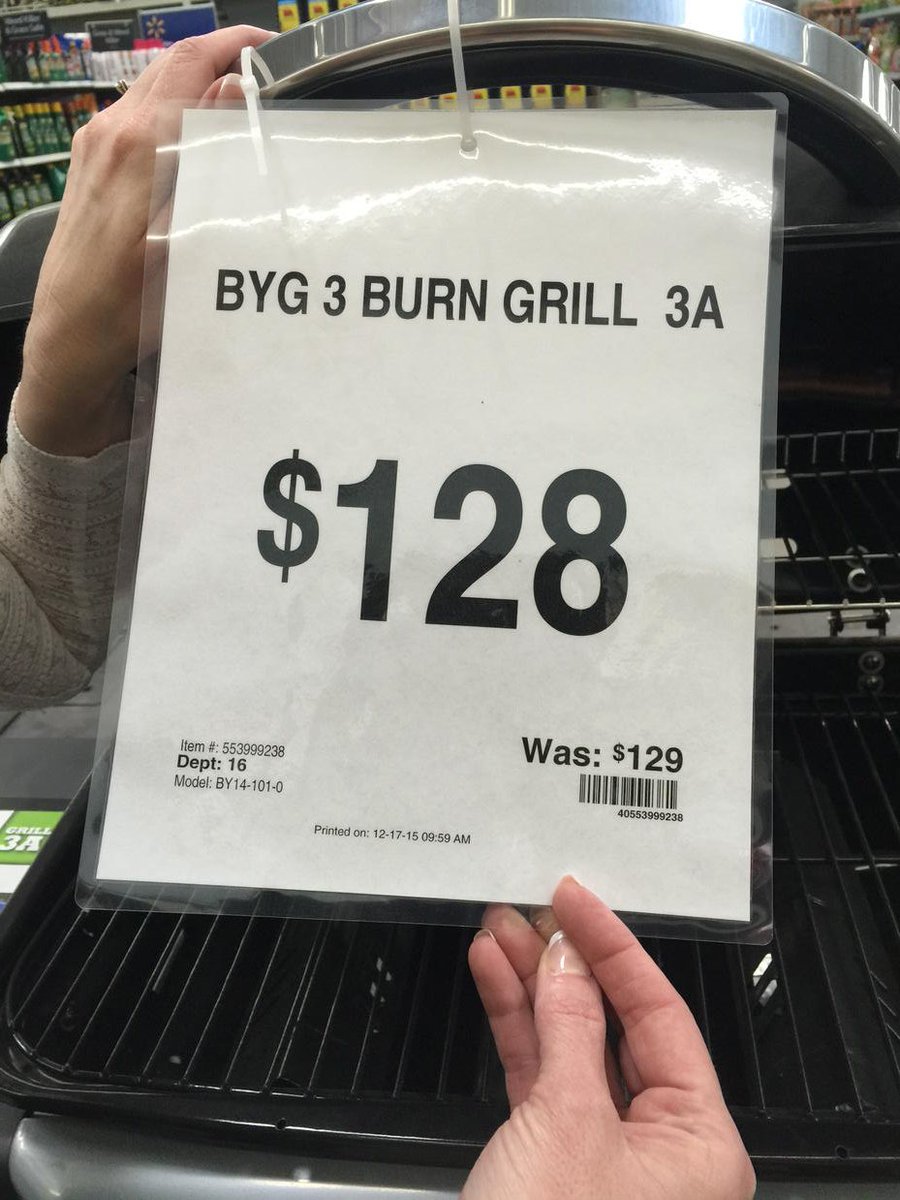} \\
 \footnotesize{(c) thanks for the awesome, leaky cup <user>, making my morning just so much better}&  \footnotesize{(d) hey <user> great sale!}
\end{tabular}
\caption{Examples of multimodal sarcasm. In examples (a) and (b), sarcasm is due to the contradiction between the text's claim and image. In (c), the image complements the sarcastic text. For (d), the image is sarcastic and the text complements the tone.} 
 \label{fig:examples}
\end{figure}

\textbf{Limitation:} Collectively, these prompt-learning methods exhibit two primary limitations.
\textbf{(a)} \textit{Unified prompt for both modalities}: They use a single prompt across both modalities. Yet, multimodal sarcasm can manifest in diverse ways: Case 1: when the image and text are incongruent (Fig. \ref{fig:examples}(a, b)), Case 2: when the image itself conveys sarcasm, with the text reinforcing that meaning (Fig. \ref{fig:examples}(d)), Case 3: when the text is itself sarcastic while the image plays a supporting role (Fig. \ref{fig:examples}(c)). We argue that a single unified prompt is insufficient to capture these varied scenarios, and therefore,  modality-disentangled prompt tuning is essential.
\textbf{(b)}. \textit{Shallow Prompt Tuning}: These methods apply prompts only at the embedding layer of the pretrained language model (PLM). However, we argue that different encoder layers in a PLM capture progressively richer and more abstract representations. Therefore, incorporating prompts at deeper layers is crucial to fully exploit the model’s capacity for nuanced understanding, especially for complex tasks like sarcasm detection.


In light of these limitations, we propose to modify CLIP (Contrastive Language–Image Pretraining) \cite{Radford2021LearningTV} in the prompt tuning framework, which we call DMDP (\textbf{D}eep \textbf{M}odality-\textbf{D}isentangled \textbf{P}rompt Tuning). CLIP is a multimodal model designed to bridge the gap between natural language and visual understanding by jointly training on almost 400 million image-text pairs using contrastive learning. Hence it presents a strong multimodal foundation (prior) to understand multimodal inputs, and has shown to perform well for multimodal sarcasm detection \cite{qin-etal-2023-mmsd2}. To capture the modality-specific sarcasm signals explicitly, we disentangle the unified prompt into two separate text and visual prompts, applied to the text and visual encoders of CLIP respectively. Going a step further, to handle the first case of sarcasm, we employ a cross-modal prompt alignment strategy, ensuring that the model captures the interplay and contrast between the modalities. For the second and third cases, where one modality is inherently sarcastic while the other acts as a supporting element, we implement a weighting mechanism on the modality prompts. This helps the model give more importance to the causal modality. This enables our model to overcome the first limitation associated with unified prompts. We observe that different layers of the visual and text encoders of CLIP learn distinct features from the input. To utilize the potential of layer-wise learning, we introduce deep prompt tuning by injecting prompts into deeper layers of both the text and visual encoders. Further, we implement a layer-wise prompt sharing mechanism, where each layer's prompt incorporates information from the previous layer, enabling progressive refinement and better contextualization across the network. This takes care of the second limitation of shallow prompt-tuning.

Additionally, to mitigate the noise caused by randomly initialized prompts, we implement a gating mechanism where the prompts in the first layer are initially gated to zero and the gate values are eventually learned during training.

Extensive experiments on two public datasets demonstrate that our model, DMDP, outperforms both prompt-based and non-prompt-based methods in the few-shot setting (using only 1\% of the data). Further, in extremely low-resource scenarios such as 5-shot, 10-shot, and 20-shot settings, DMDP consistently surpasses large vision-language models (LVLMs) by significant margins, while utilizing over 13 times fewer parameters. Notably, our model also achieves superior performance in out-of-distribution (OOD) evaluation, highlighting its strong generalization capabilities.

\textbf{Contributions}: The main contributions of our work are:
\begin{itemize}

\item We propose DMDP, a novel framework with deep continuous prompts across CLIP’s text and image encoders, featuring layer-wise prompt sharing, cross-modal prompt alignment, and a modality-weighting strategy.

\item We identify three distinct types of multimodal sarcasm, which were not explicitly handled by the prior approach \texttt{CAMP}, and propose a modality-disentangled prompt-tuning strategy to address them.

\item We perform extensive experiments on two public datasets and demonstrate our model's superiority in few-shot and OOD settings.
\end{itemize}

\section{Related Work}

\subsection{Image-Text Sarcasm Detection}
Traditional sarcasm detection methods relied on text modality only. Early approaches used lexical inconsistencies \cite{Joshi2015HarnessingCI,khattri-etal-2015-sentiment, joshi-etal-2016-word,amir-etal-2016-modelling} and sequence modeling approaches \cite{zhang-etal-2016-tweet, poria-etal-2016-deeper, ghosh-etal-2017-role, Agrawal2018AffectiveRF, Agrawal2020LeveragingTO, babanejad-etal-2020-affective, Lou2021AffectiveDG, liu-etal-2022-dual} to detect textual sarcasm. \citet{Schifanella2016DetectingSI} used hand-crafted image-text features and paved the way for multimodal image-text sarcasm detection. \citet{cai-etal-2019-multi} proposed a hierarchical fusion model that integrates text, image features, and image attributes to improve multimodal sarcasm detection on Twitter. \citet{xu-etal-2020-reasoning} proposed a model that captures cross-modality contrast and semantic associations to improve reasoning in multimodal sarcastic tweets. \citet{pan-etal-2020-modeling} demonstrated that sarcasm can stem from intra-modal and inter-modal associations, leading them to propose a self-attention model to capture incongruities within and between modalities. To detect fine-grained incongruity between image and text tokens, \citet{Liang2021MultiModalSD, Liang2022MultiModalSD} employed graph neural networks on image and text token graphs. \citet{liu-etal-2022-towards-multi-modal} introduced a hierarchical framework incorporating external knowledge to enhance multimodal sarcasm detection. \citet{tian-etal-2023-dynamic} proposed a model that adaptively captures inter-modal contrasts between image and text using dynamic routing mechanisms to improve sarcasm detection. \citet{qin-etal-2023-mmsd2} proposed a cleaner version of MMSD dataset and builds on the CLIP framework by incorporating perspectives from image, text, and image-text interactions to enhance multimodal sarcasm detection. Work by \citet{Tang2024LeveragingGL} used retrieval augmentation with instruction tuning for effective sarcasm detection. 

Compared to previous works, we utilize a prompt tuning paradigm with a frozen vision-language model to effectively use the pre-trained knowledge to minimize the risk of overfitting on few-shot multimodal sarcasm detection.

\begin{figure*}[t]
\centering
\label{fig:pemcamp}
\includegraphics[width=0.98\textwidth, keepaspectratio]{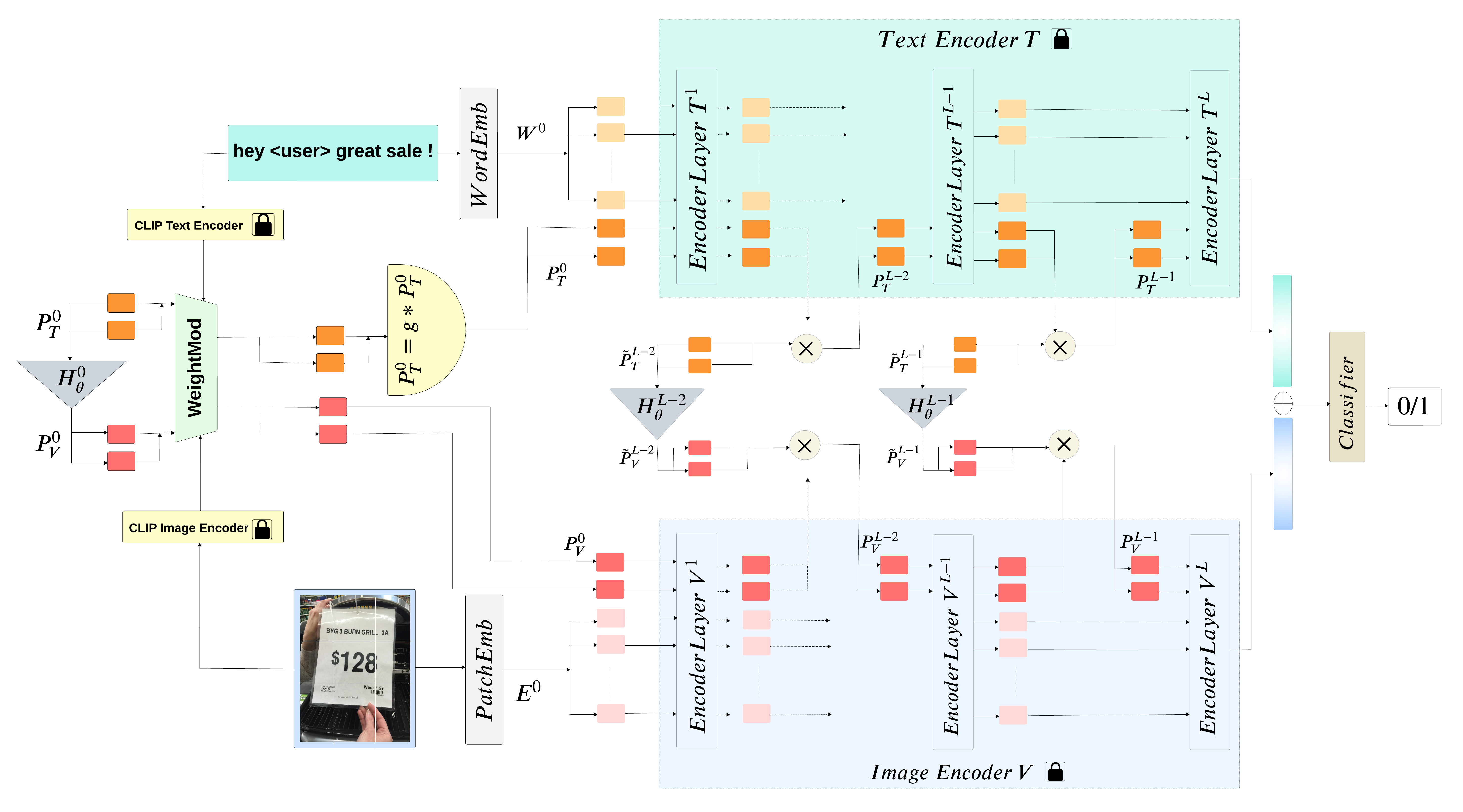} 
\caption{\label{fig:model}Architecture of our DMDP model.}
\end{figure*}

\subsection{Prompt Tuning}
\citet{Li2021PrefixTuningOC} proposed the idea of prompt tuning in NLP. They used learnable continuous tokens instead of manually handcrafted prompts for table-to-text generation and summarization tasks. \citet{Lester2021ThePO} simplified their approach by applying prompts only in the embedding layer. \citet{Jia2022VisualPT} first introduced prompt learning for the vision transformer model. A series of works \cite{Yoo2023ImprovingVP, Han2023E2VPTAE, Mo2024LSPTLS} improved visual prompt tuning using the ViT model. \citet{Zhou2021LearningTP} proposed prompt tuning in the multimodal CLIP. \citet{Zhou2022ConditionalPL} improved upon that work by incorporating image-conditioned prompts. \citet{Hegde2023CLIPG3} used prompt tuning CLIP for language guided 3D recognition. \cite{Wasim2023VitaCLIPVA, Wang2024ViLTCLIPVA} used multimodal prompt-learning in CLIP for video understanding. \cite{Wang2023SeeingIF, Xing2023MultimodalAO} adapted CLIP under prompt tuning for action recognition.
\citet{Jana2024ContinuousAM} proposed a model, CAMP, that used a BERT backbone with attentive continuous prompt-tuning for few-shot sarcasm detection.

However, all of these methods use an unimodal prompt tuning approach, which fails to capture subtle interactions between the modalities. To this end, our proposed approach uses modality-disentangled prompt tuning for few-shot multimodal sarcasm detection.

\section{Preliminaries}
In this section, we revisit the working of CLIP (Contrastive Language–Image Pretraining), since we propose to modify CLIP. It is a foundational vision-language model that consists of a vision encoder and a text encoder. Below, we summarize the key components of CLIP: \\  \\
\textbf{Vision Encoder:} Image $I$ is first divided into m fixed-sized patches. Each patch is projected into $d_{v}$ dimensional embedding space with positional embeddings attached to each patch token.
\setlength{\abovedisplayskip}{5pt}  
\begin{equation}
E^{0} = \text{PatchEmb}(I) + \text{Pos}
\quad E^{0} \in \mathbb{R}^{m \times d_{v}}
\end{equation}
where $E^{0}$ is the set of patch embeddings for the first transformer layer $V^{1}$.
The set of patch embeddings $E^{i}$ are prepended with a learnable class embedding token $z_{i}$ and passed through L transformer blocks as follows:
\begin{equation}
[z^{i}, E^{i}] = \text{V}^{i}([z^{i-1}, E^{i-1}])
\quad i = 1, \dots, L
\end{equation}
The final image embedding $i$ is obtained by projecting $z^{L}$ into a shared embedding space as follows:
\begin{equation}
i = \text{Proj}_{I}(z^{L})
\quad i \in \mathbb{R}^{d}
\end{equation}
\textbf{Text Encoder:} Given an input text sequence $l$, it is tokenized into n tokens $[l_{1}, l_{2},.., l_{n}]$. The tokens are converted to $d_{t}$ dimensional word embeddings with positional embeddings added.
\begin{equation}
W^{0} = \text{WordEmb}(l) + \text{Pos}
\quad W^{0} \in \mathbb{R}^{n \times d_{t}}
\end{equation}
Subsequently, they are processed sequentially through L transformer blocks as follows:
\begin{equation}
[W^{i}] = \text{T}^{i}([W^{i-1}])
\quad i = 1, \dots, L
\end{equation}
The final text embedding $t$ is obtained by projecting the embedding of the last token of the output from the last transformer layer into a shared embedding space as:
\begin{equation}
t = \text{Proj}_{T}(W^{L}[-1])
\quad t \in \mathbb{R}^{d}
\end{equation}

\section{Methodology}

\subsection{Problem Definition}
Given a multimodal input sample $x_{j} = (T_{j}, I_{j})$, where 
\begin{equation}
T_{j} = \{t^{1}_{j}, t^{2}_{j}, ...., t^{n}_{j} \}
\end{equation}is the textual content and $I_{j}$ is the associated image, the objective is to classify $x_{j}$ with a label $y_{j}$ from the set $Y = \{sarcastic, nonsarcastic \}$.

\subsection{Deep Modality-Disentangled Prompt Tuning}
Figure \ref{fig:model} shows the overall architecture of our proposed model DMDP.
Sarcasm is a complex phenomenon, thus, a single unified prompt is not sufficient to capture modality-specific signals. Hence, we disentangle the prompts and apply them to the visual and text encoders separately, with a cross-modal prompt alignment strategy to foster cross-modal interaction. Further, to learn distinct signals from each layers of the encoders, we incorporate these prompts into deeper layers of the encoders with a layer-wise prompt sharing strategy. This process is explained below.

\subsubsection{Text Modality Prompt Tuning}
To capture sarcastic signals from the text, we append learnable prompt tokens $P_{T}$ to the input embedding of the text encoder of CLIP, where $P^{T} \in \mathbb{R}^{c \times d_{t}}$. The text encoder can now be reformulated as:
\begin{equation}
[P_{T}^{i}, W^{i}] = \text{T}^{i}([P_{T}^{i-1}, W^{i-1}])
\quad i = 1,.., S
\end{equation}
To capture the distinct information at each layer, we introduce new learnable tokens $\tilde{P}_{T}$ up to S transformer layers, where $S \leq L$. Directly introducing new learnable tokens in every layer may lead to information loss from the previous layer and will lead to unstable training. To mitigate these issues, we introduce a prompt sharing mechanism, where the current layer's prompt tokens share the information from the previous layer. Thus, the prompt tokens to be fed to transformer layer $i$ are computed as:
\begin{equation}
   P_{T}^{i-1} =  P_{T}^{i-1} * \tilde{P}_{T}^{i-1}
   \quad 1 < i \leq S
\end{equation}
This adaptive scaling mechanism allows the model to weigh how much influence the previous layer’s information should have in the current layer.
The process of extracting the final text representation remains the same as that of the original CLIP.
\begin{equation}
t = \text{Proj}_{T}(W^{L}[-1])
\quad t \in \mathbb{R}^{d}
\end{equation}

\subsubsection{Visual Modality Prompt Tuning}
To extract sarcasm-specific features from the image, similar to text prompting, we incorporate prompt tokens $P_{V}$ to the input embedding of the vision encoder, where $P_{V} \in \mathbb{R}^{c \times d_{v}}$. 
Independent learning of vision and text prompt tokens will not capture the sarcastic interactions between the two, needed to detect cases illustrated in Fig. \ref{fig:examples}(a and b). To overcome this limitation, we design a linear mapping function that projects all the textual prompt tokens to visual tokens to achieve more coherence in the learning process. \textbf{We also experimented by constructing textual prompt tokens as projections of visual tokens, which yield inferior results.} The details are in Analysis subsection \ref{subsec:reverse}.
\begin{multline}
P_{V}^{i} = H_{\theta}^{i}(P_{T}^{i})  \quad H_{\theta}^{i} \in \mathbb{R}^{d_{t} \times d_{v}} \\
\hspace*{1cm} 0 \leq i \leq S
\end{multline}
The vision encoder can now be reformulated as:
\begin{multline}
[z^{i}, E^{i}, P_{V}^{i}] = \text{V}^{i}([z^{i-1}, E^{i-1}, P_{V}^{i-1}]) \\
\hspace*{1cm} i = 1, \dots, S \leq L
\end{multline}
We apply deep layer prompting similar to text encoder up to $S$ transformer layers. The prompt tokens to be fed to transformer layer $i$ are computed as:
\begin{equation}
   P_{V}^{i-1} =  P_{V}^{i-1} * \tilde{P}_{V}^{i-1}
   \quad 1 < i \leq S
\end{equation}
The final image embedding $i$ is obtained by projecting $z^{L}$ into a shared embedding space as follows:
\begin{equation}
i = \text{Proj}_{I}(z^{L})
\quad i \in \mathbb{R}^{d}
\end{equation}

\subsection{Modality-Weighted Prompts}
We introduce a weighting mechanism for the modality prompts to cater to cases where one modality carries a sarcastic tone and the other serves a supporting role. This enables the model to assign greater significance to the relevant modality, facilitating a more adaptable and precise understanding of the driving modality. To achieve this, we implement a module named WeightMod to get the weighted coefficient as follows:
\begin{equation}
r = \text{WeightMod}([\tilde{i}, \tilde{t}]))  
\end{equation}
where WeightMod is implemented as a single linear layer with a sigmoid non-linearity. $\tilde{i}$ and $\tilde{t}$ are image and text embeddings from the original CLIP model without using any prompt learning. This weighting coefficient is applied to both the modality prompts only for the first layer to determine their importance.
\begin{align}
P_{T}^{0} &= r * P_{T}^{0} \\
P_{V}^{0} &= (1 - r) * P_{V}^{0}
\end{align}

\begingroup
\begin{table*}[t]
\caption{\label{baselines}
Statistics of few-shot MMSD and MMSD2.0 dataset provided by \cite{Jana2024ContinuousAM}. For splits presented as X / Y, X represents the few-shot data sampled while Y represents the total data. The total train split represents approximately 1\% of the total training data with $|valid| = |train|$, while the number of samples in the test set is kept the same.
}
\label{tab:dataset}
\centering
\renewcommand{\arraystretch}{0.9}
\scriptsize
\begin{adjustbox}{width= 1.9\columnwidth}
\begin{tabular}{lcccccccccc}
\toprule
 \multirow{2}{*}{\textbf{Dataset}}& \multicolumn{3}{c}{\textbf{Train}}
 & \multicolumn{3}{c}{\textbf{Valid}}
 & \multicolumn{3}{c}{\textbf{Test}} \\

\cmidrule(lr){2-4} \cmidrule(lr){5-7} \cmidrule(lr){8-10}
& \textbf{Pos} & \textbf{Neg} & \textbf{Total} & \textbf{Pos} & \textbf{Neg} &  \textbf{Total} & \textbf{Pos} & \textbf{Neg} & \textbf{Total} \\
\midrule

 MMSD& 99  / 8642 & 99 / 11174 & 198 / 19816& 99 / 959& 99 / 1459& 198 / 2410 & 959 / 959 & 1450 / 1450& 2409 / 2409&\\
\midrule
 
 MMSD2.0 & 99 / 9576 & 99 / 10240 & 198 / 19816 & 99 / 1042 & 99 / 1368 & 198 / 2410 & 1037 / 1037 & 1072 / 1072 & 2409 / 2409 &\\

\bottomrule
\end{tabular}
\end{adjustbox}

\end{table*}
\endgroup

\subsection{Gated Prompt Tuning}
Learnable prompt tokens, introduced in the transformer layers, are randomly initialized. This results in noise being introduced to the pre-trained CLIP encoders during the initial training phase and hence results in degraded performance. To curb this, we incorporate a parameterized gating mechanism to the textual prompt tokens in the first layer. 
\begin{equation}
g = \frac{e^{\tau} - e^{-\tau}}{e^{\tau} + e^{-\tau}}
\end{equation}
$\tau$ is the gate prior and is initially set to 0.5.
\begin{equation}
 P_{T}^{0} = g * P_{T}^{0}   
\end{equation}
Since subsequent text prompt tokens are weighted by the previous tokens, gating the first token is sufficient to curb the noise. The image prompt tokens are generated as a projection of the text prompt tokens. This projection ensures that the image tokens inherently inherit the characteristics of the gated text tokens. Since $P_{T}^{0}$ has already been stabilized through the gating mechanism, the image tokens indirectly benefit from this stabilization.

\subsection{Model Training and Prediction}
The final image representation $i$ and text representation $d$ are extracted from the model. They are concatenated and passed through a classification layer $f_{\gamma}$ for the final prediction. During training, textual prompt tokens $P_{T}^{0\, \text{to}\, S-1}$, visual prompt tokens $P_{V}^{0\, \text{to}\, S-1}$
, mapping layer $H_{\theta}$, modality weighting layer WeightMod, the gating module and the final classifier $f_{\gamma}$ are updated using cross-entropy loss. CLIP's pre-trained weights are frozen throughout the training process. 

\section{Experiments}

\subsection{Datasets}
We evaluate our model DMDP on the few-shot split of MMSD \cite{cai-etal-2019-multi} and MMSD2.0 \cite{qin-etal-2023-mmsd2}, provided by \cite{Jana2024ContinuousAM}. In line with the few-shot setting described in \cite{Yu2022FewShotMS, Yang2022FewshotMS, Yu2022UnifiedMP}, \cite{Jana2024ContinuousAM} provides two few-shot splits (1\% of the training set) for each dataset, ensuring an equal number of samples per category. They set the validation set size equal to the training set size ($|valid| = |train|$), while keeping the test set size same as the original. Dataset statistics are shown in Table \ref{tab:dataset}.

\subsection{Experimental Settings}
We use ViT-B/16 CLIP model in a prompt tuning framework. We set the prompt length $c$ to 2 and prompt depth $S$ to 9 for all the experiments. Few-shot training shows quite a variation in performance. To account for this, for each dataset, we perform 6 runs (3 times x 2 splits) and report the mean Accuracy (Acc), mean Macro-F1 (F1), and the standard deviation across the 6 runs. Our model is trained with a batch size of 4, learning rate of 0.0035, warmup epoch of 1 and warmup learning rate of 1e-5. We train our model for 100 epochs and select the model with the best performance on the validation set for final testing. We run all our experiments on a Nvidia RTX A5000 GPU with 24GB of memory.

\begingroup

\begin{table}[t]
\centering
\small
\renewcommand{\arraystretch}{1.3}
\setlength{\tabcolsep}{4pt}
\caption{\label{baselines}
Performance comparison of existing methods with our proposed model DMDP in few-shot setting. The best results across metrics are highlighted in bold. Numbers in brackets indicate standard deviation. Our method outperforms baselines significantly with p < 0.05.}
\label{tab:main_results}
\begin{adjustbox}{max width=0.9\columnwidth}
\begin{tabular}{l l l l l l}
\toprule
& & \multicolumn{2}{c}{\textbf{MMSD}}
 & \multicolumn{2}{c}{\textbf{MMSD2.0}} \\
\midrule
\textbf{Modality} & \multicolumn{1}{l}{\textbf{Method}} & \textbf{Acc} & \textbf{F1} & \textbf{Acc} & \textbf{F1}\\
\midrule
 
 & ResNet \cite{He2015DeepRL} & 0.664 (0.1) & 0.602 (1.2) & 0.638 (1.3) & 0.625 (0.5)\\
Image & ViT \cite{Dosovitskiy2020AnII}  & 0.611 (1.6) & 0.522 (1.7) & 0.560 (2.8) & 0.614 (0.5)\\
 & VPT-S \cite{Jia2022VisualPT} & 0.634 (0.8) & 0.624 (0.4) & 0.618 (1.4) & 0.617 (1.4)\\
& VPT-D \cite{Jia2022VisualPT} & 0.641 (2.1) & 0.636 (1.8) & 0.634 (1.1) & 0.631 (1.1)\\

\midrule
 & TextCNN \cite{Kim2014ConvolutionalNN} & 0.631 (2.8) & 0.549 (2.5) &0.568 (0.7) & 0.570 (1.6)\\
& BiLSTM \cite{Graves20052005SI}  & 0.602 (1.7) & 0.560 (2.3) & 0.499 (2.1) & 0.595 (2.1)\\
Text  & RoBERTa \cite{Liu2019RoBERTaAR} & 0.689 (2.2) & 0.675 (3.1) & 0.598 (2.9) & 0.591 (2.4)\\
& LM-BFF \cite{gao-etal-2021-making} & 0.695 (2.7) & 0.688 (2.3) & 0.637 (1.4) & 0.626 (2.5) \\
& LM-SC \cite{jian-etal-2022-contrastive} & 0.698 (1.4) & 0.681 (0.8) & 0.640 (0.7) & 0.632 (1.5)\\

\midrule
 & HFM \cite{cai-etal-2019-multi}  & 0.612 (1.3) & 0.598 (1.1) & 0.561 (0.2) & 0.361 (0.3)\\
& Att-BERT \cite{pan-etal-2020-modeling} & 0.707 (1.7) & 0.696 (1.3) & 0.659 (1.6) & 0.683 (1.8)\\
 Image+Text& HKE \cite{liu-etal-2022-towards-multi-modal}  & 0.503 (2.3) & 0.667 (2.8) & 0.408 (1.5) & 0.579 (1.3) \\
(Non-Prompt-Based)& DIP \cite{Wen2023DIPDI} & 0.704 (2.7) & 0.698 (2.3) & 0.685 (2.8) & 0.658 (2.6) \\
& MV-CLIP \cite{qin-etal-2023-mmsd2}  & 0.780 (0.2) & 0.770 (0.2) & 0.742 (0.4) & 0.740 (0.3) \\
& DynRT \cite{tian-etal-2023-dynamic}  & 0.583 (0.1) & 0.487 (0.6) & 0.518 (2.9) & 0.513 (3.2) \\
& RAG-LLaVA \cite{Tang2024LeveragingGL}  & 0.483 (9.1) & 0.406 (4.6) & 0.569 (0.1) & 0.446 (8.3) \\

\midrule
 & PVLM \cite{Yu2022FewShotMS}  & 0.712 (0.6) & 0.699 (0.2) & 0.665 (2.2) & 0.658 (2.1) \\
Image+Text& UP-MPF \cite{Yu2022UnifiedMP} & 0.707 (2.4) & 0.701 (2.6) & 0.669 (0.4) & 0.663 (0.1) \\

(Prompt-Based)&CoOp \cite{Zhou2021LearningTP} & 0.772 (1.6) & 0.769 (1.5) & 0.759 (0.9) & 0.759 (0.8) \\

&CoCoOp \cite{Zhou2022ConditionalPL} & 0.782 (1.0) & 0.779 (1.0) & 0.746 (1.6) & 0.745 (1.6)\\

&CAMP \cite{Jana2024ContinuousAM} & 0.729 (0.9) & 0.717 (1.0) & 0.692 (2.8) & 0.681 (2.3)\\
\midrule
&DMDP (ours) & \textbf{0.810 (0.3)} & \textbf{0.806 (0.4)} & \textbf{0.775 (0.5)} & \textbf{0.774 (0.3)}\\
\bottomrule
\end{tabular}
\end{adjustbox}
\end{table}
\endgroup

   \begin{figure*}[t]
    \centering
    \label{fig:general_model}
    \includegraphics[width=0.96\textwidth, keepaspectratio]{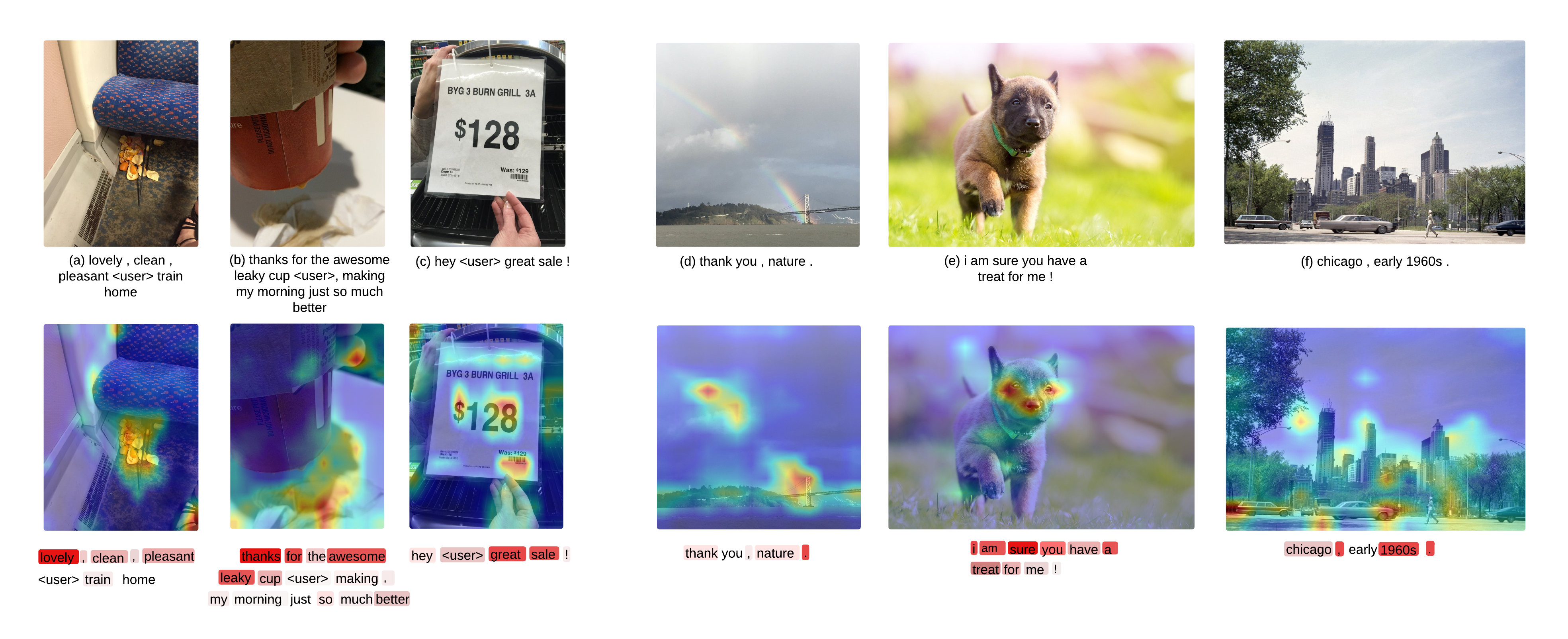} 
    \caption{\label{fig:heat}Attention visualizations of textual and visual prompt tokens over the text and image respectively from the last transformer layer. The first row shows the original image and text. [a, b, c] indicate sarcastic samples and [d, e, f] indicate non-sarcastic samples.}
    \end{figure*}

\section{Baselines}
We evaluate our model on four groups of baselines, namely, image-modality, text-modality, non-prompt-based multimodal and prompt-based multimodal methods. For the image-modality baselines, we consider \textbf{ResNet} \cite{He2015DeepRL}, \textbf{ViT} \cite{Dosovitskiy2020AnII}, and \textbf{VPT-S}; \textbf{VPT-D} \cite{Jia2022VisualPT} which used prompt tuning in ViT in the shallow and deep layers, respectively. 

For the text-modality baselines, we compare with \textbf{TextCNN} \cite{Kim2014ConvolutionalNN}, \textbf{Bi-LSTM} \cite{Graves20052005SI}, \textbf{RoBERTa}, \cite{Liu2019RoBERTaAR}, \textbf{LM-BFF} \cite{gao-etal-2021-making}, which utilized hard prompt-based finetuning for pre-trained language models and,  \textbf{LM-SC} \cite{jian-etal-2022-contrastive} which improved LM-BFF by employing contrastive learning on the text.

For the non-prompt-based multimodal baselines for sarcasm detection, we consider \textbf{HFM} \cite{cai-etal-2019-multi} which used hierarchical early and late fusion to fuse image, text and image attributes, \textbf{D\&R Net} \cite{xu-etal-2020-reasoning} which employed decomposition and relation network to determine semantic association, \textbf{Att-BERT} \cite{pan-etal-2020-modeling} which used co-attention networks to detect inter and intra-modal incongruity, \textbf{InCrossMGs} \cite{Liang2021MultiModalSD} which employed in-modal and cross-modal graphs to model incongruity, \textbf{CMGCN} \cite{Liang2022MultiModalSD} which used a cross-modal image-text weighted graph, and \textbf{HKE} \cite{liu-etal-2022-towards-multi-modal} that modeled coarse and fine-grained incongruities using a hierarchical interaction network. \textbf{MV-CLIP} \cite{qin-etal-2023-mmsd2} used CLIP with an image-text interaction layer to capture cross-modal incongruity, \textbf{DIP} \cite{Wen2023DIPDI} captured sarcasm by modeling incongruity at both factual and affective levels through semantic reweighting, uncertainty modeling, and contrastive learning, \textbf{DynRT} \cite{tian-etal-2023-dynamic} which used a dynamic routing mechanism using transformer model to determine the tokens responsible for sarcasm and \textbf{RAG-LLaVA} \cite{Tang2024LeveragingGL} that utilized similarity-based retrieval module to fetch demonstration examples to assist LLaVA for sarcasm label generation.

In the prompt-based multimodal baselines, we consider, \textbf{PVLM} \cite{Yu2022FewShotMS} that used prompt fine-tuning while incorporating image tokens, \textbf{UP-MPF} \cite{Yu2022UnifiedMP} which pretrained initially on image tasks to narrow the semantic gap between image and textual prompts before applying it to sentiment analysis. Since we use prompt-learning in CLIP, we also consider two CLIP-based adaptations used for image recognition; \textbf{CoOp} \cite{Zhou2021LearningTP}, which introduces prompt tokens only in the text encoder of CLIP, \textbf{CoCoOp} \cite{Zhou2022ConditionalPL} that uses image-conditioned text prompts. Finally, we also consider the only baseline for few-shot multimodal sarcasm detection task \textbf{CAMP} \cite{Jana2024ContinuousAM}, which used continuous attentive prompt tokens.

We evaluate all baseline models on the few-shot data splits using their original setting and present the results. Some baselines are excluded from comparison due to the unavailability of their codebase.\footnote{The source codes for D\&R Net and InCrossMGs are not publicly accessible, and CMGCN relies on additional attributes that are not available.}

\section{Main Results}
Following \cite{Jana2024ContinuousAM}, we report the few-shot results in Table \ref{tab:main_results}. Our findings are as follows: (1) Our method DMDP outperforms all SoTA baselines (Acc: $\uparrow 2.8\%$ in MMSD and $\uparrow 1.6\%$ in MMSD2.0) across both unimodal and multimodal settings. 
(2) All the prompt-based methods perform comparable or even better than their non-prompt-based counterparts. Since these models are frozen during training and only the continuous learnable prompts get updated, the model can preserve its pre-training knowledge and avoid catastrophic forgetting. 
(3) Among the prompt-based models, CLIP-based adaptations: CoOp, CoCoOp, and DMDP show significant improvements over other non-CLIP-based methods. This demonstrates the impact of the strong multimodal prior that CLIP provides.
(4) Our model DMDP, which employs prompt tokens in both image and text encoders, significantly outperforms CoOp and CoCoOp, which use prompt tokens in the text encoder only. This shows the importance of modality-disentangled prompt tuning as it enables the shared learning of knowledge across both modalities, which is crucial to deciphering the subtle image-text interactions necessary for sarcasm. Further, our gating and prompt weighting strategies dynamically adjust the contribution of each modality, ensuring that the model focuses on the most relevant features from both the image and text.






\section{Comparison with LVLMs in Low-Resource Settings}
\begin{figure}
\centering
\includegraphics[width=0.85\columnwidth]{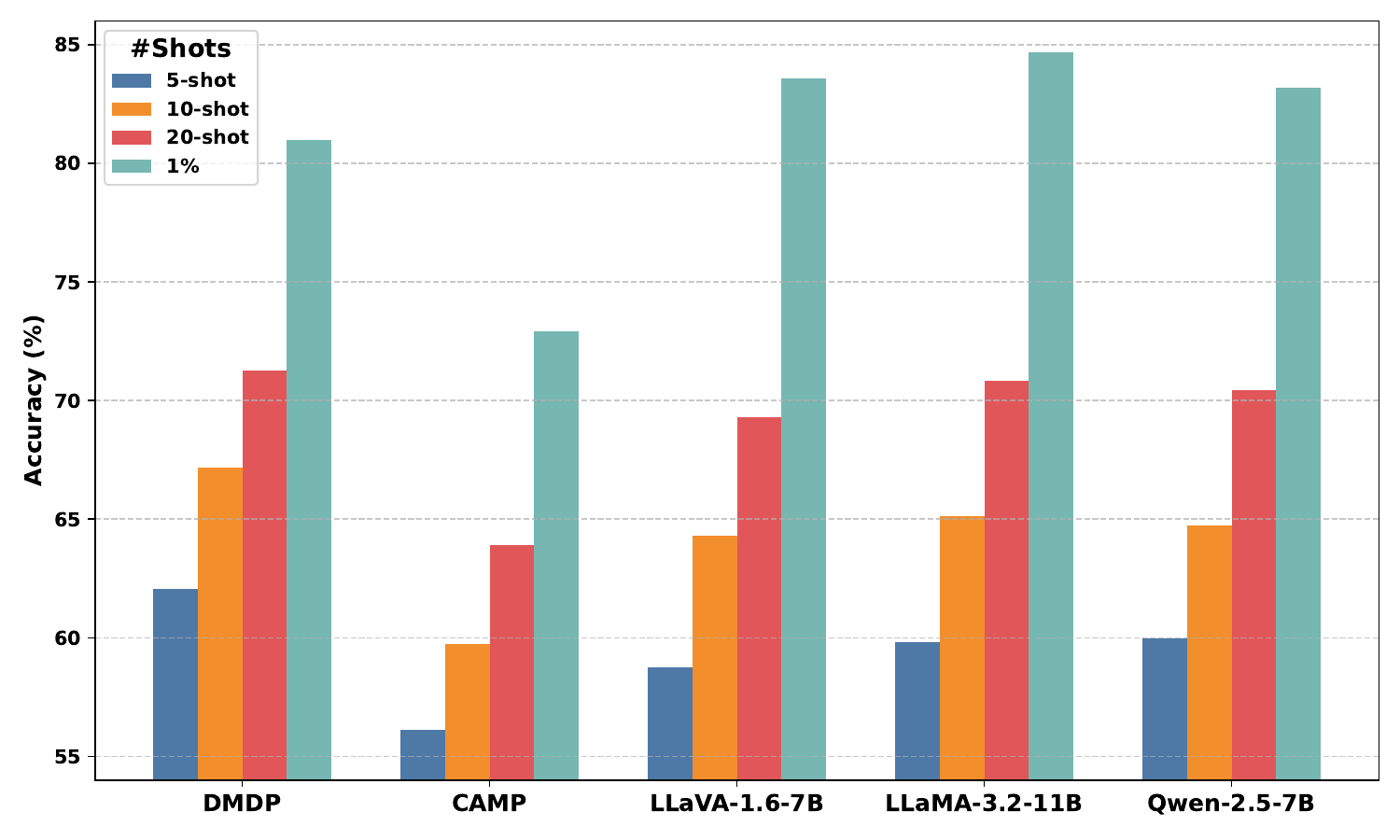} 
\caption{\label{fig:mmsd_low_res}Performance comparison for low-resource MMSD dataset.}
\end{figure}

\begin{figure}
\centering
\includegraphics[width=0.85\columnwidth]{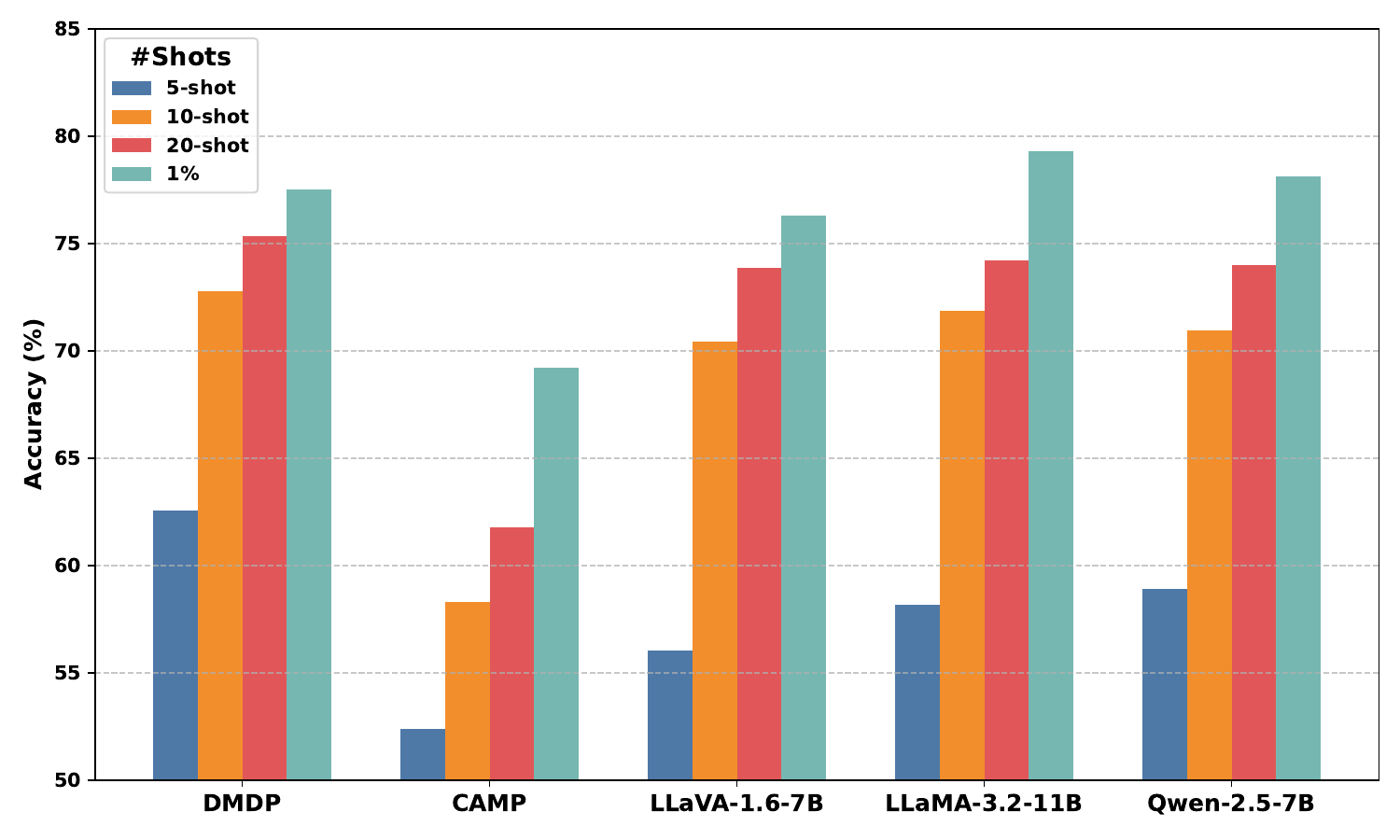} 
\caption{\label{fig:mmsd2_low_res}Performance comparison for low-resource MMSD2.0 dataset.}
\end{figure}

To further assess the strength of our model, we conducted a series of experiments with extremely low-resource scenarios using 5-shot, 10-shot and 20-shot settings on both MMSD and MMSD2.0 datasets. From Figures \ref{fig:mmsd_low_res} and \ref{fig:mmsd2_low_res}, we observe that our model outperforms \texttt{CAMP} and other LVLMs in these settings, with the performance gap being significant as the number of shots decreases. Compared to large-scale LVLMs, which typically rely on massive parameter counts and extensive pre-training data, our approach offers a more efficient and adaptable alternative. While LVLMs often struggle to generalize in low-shot scenarios due to their dependence on scale and data quantity, our modality-disentangled deep prompt-tuning framework achieves superior performance with significantly fewer trainable parameters. Notably, even in the 1\% data setting, DMDP exhibits near-comparable accuracy, with over 13 $\times$ fewer trainable parameters, highlighting its efficiency and practicality in resource-constrained environments.

\section{Analysis}
To further prove the efficacy of DMDP, we answer the following research questions:
(1) \textit{What do the visual and textual prompt tokens learn?} 
(2) \textit{What is the effect of disentangling modality prompts?} 
(3) \textit{What is the implication of deep vs shallow prompt tuning?} 
(4) \textit{What effect does gating have on prompt learning? } 
(5) \textit{What is the impact of dynamic weighting on prompts?} 
(6) \textit{What is the effect of removing the cross-modal prompt alignment?} 
(7) \textit{How do textual prompts constructed as a projection of visual prompts perform?}
(8) \textit{What is the implication of prompt length?}

\subsection{Attention Visualization of Prompt Tokens}
Figure \ref{fig:heat} visualizes the attention distribution of the visual and textual prompt tokens from the final transformer layers over the image and text, respectively. The first three examples [a, b, c] represent sarcastic cases, while the latter three [d, e, f] are non-sarcastic. In (a), where sarcasm is due to the contradiction between the text and the image, the textual prompts focus on positive words like \textit{thanks}, \textit{clean}, and \textit{pleasant}, while the visual prompts concentrate on the orange peels, indicating dirt, allowing the model to detect incongruity despite the text not mentioning the orange peel. In (b), the sarcasm arises from the user’s ironic gratitude for a leaky cup. Here the text modality is itself sufficient to indicate sarcasm and the image modality reinforces the sarcastic undertone. The text tokens attend to sarcastic cues like \textit{thanks}, \textit{awesome}, and \textit{leaky cup} indicating the incongruity within the text itself. The visual tokens accurately highlight the leak. In (c), the sarcasm arises from the exaggerated praise of a discount of only \$1 being called a great sale. This is the case of the image portraying sarcasm, and the text highlighting the sarcastic sense. The visual prompts successfully attend to two key regions in the image: the price tag showing \$128 and another showing \$129, capturing the contrast from the image itself. Simultaneously, the textual prompts focus on the phrase \textit{great sale}, enabling the model to capture the discrepancy between the text's exaggerated positivity and the trivial nature of the discount.

For non-sarcastic instances, while the visual prompts still highlight relevant parts of the image, the textual prompts show a more balanced attention distribution across the text. In example (d), despite the presence of a potential sarcasm trigger word like \textit{thank you}, the textual prompts do not primarily focus on it, as the context lacks a sarcastic undertone.

\subsection{Strength of Modality-Disentangled Prompts}
We demonstrate the weakeness of having a unified prompt for both visual and text modalities (as in \texttt{CAMP}) by employing a shared prompt across both the encoder branches. We notice a significant drop in performance (Acc: $\downarrow 2.7\%$ in MMSD and $\downarrow 2.8\%$ in MMSD2.0), as reported in Table \ref{tab:ablation} as \texttt{w UPT} (row 2). This observation highlights the importance of modality-disentangled prompts, where each modality is allowed to capture and encode its own modality-specific patterns without interference. It confirms that forcing a shared prompt hampers the model’s ability to effectively attend to complementary but distinct modality signals.

\subsection{Impact of Deep Prompt Tuning}
\begin{figure}[t]
  
\includegraphics[width=0.43\textwidth, height=0.23\textheight]{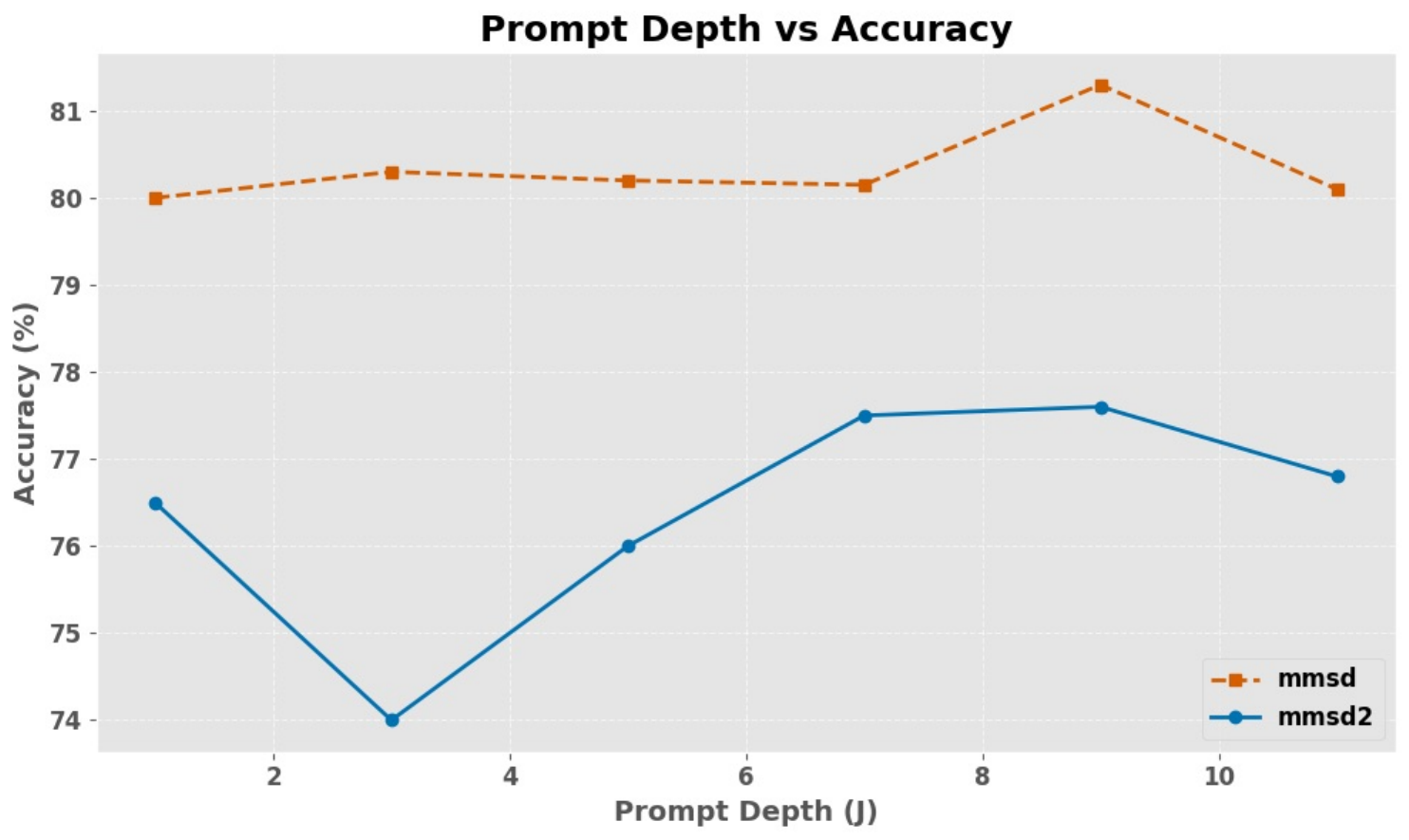}
\caption{Effect of Prompt Depth}
\label{fig:depth}
\end{figure}

We ablate on the effect of inserting prompt tokens into the deeper layers of both the visual and text encoders. From Figure \ref{fig:depth}, we observe that the performance increases as the depth of prompt tokens increases across both datasets. This trend is followed up to layer 9 (out of 12 layers). We reason that different encoder layers capture distinct types of information, with earlier layers focusing on low-level features and later layers integrating more complex patterns. The final few layers are crucial for synthesizing and combining the learned information from earlier layers. Introducing new tokens at these later stages disrupts this integration process, leading to a degradation in performance.

\subsection{Prompt Gating}
To understand the significance of the gating mechanism, we evaluate DMDP without it. From Table \ref{tab:ablation}, we observe a noticeable drop in performance (Acc: $\downarrow 0.7\%$ in MMSD and $\downarrow 0.7\%$ in MMSD2.0) for the variant \texttt{w/o Gating} (row 4). This decline highlights the critical role of gating in integrating prompt tokens effectively into the pre-trained CLIP architecture. Introducing prompt tokens directly, without gating, disrupts the pre-trained weights of CLIP by introducing noise during the early stages of learning. This disruption negatively impacts the model's ability to retain the rich multimodal prior encoded within CLIP, thereby impairing its overall performance. The gating mechanism, which starts with an initial value of 0, mitigates this issue by ensuring a gradual and controlled integration of the new prompt tokens.

\subsection{Modality-Weighted Prompts}
To assess the impact of using modality-weighted prompts, we ablate DMDP by removing the WeightMod component. From the results presented in Table \ref{tab:ablation}, we observe that the variant \texttt{DMDP w/o WeightMod} (row 5) exhibits a noticeable performance drop (Acc: $\downarrow 1.6\%$ in MMSD $\downarrow 1.4\%$ in MMSD2.0). This performance decline highlights the critical role of modality weighting in scenarios where sarcasm detection relies on a balanced understanding of both visual and textual cues. 

\begingroup
\begin{table}[t]
\caption{\label{tab:ablation} Ablation experiments of our model DMDP. Numbers indicated in brackets denote standard deviation. w indicates with that component while w/o indicates without that component. 
}

\centering
\renewcommand{\arraystretch}{1.2} 
\setlength\tabcolsep{5pt} 
\scriptsize 
\begin{tabular}{l c c c c}
\toprule
 & \multicolumn{2}{c}{\textbf{MMSD}} & \multicolumn{2}{c}{\textbf{MMSD2.0}} \\
\midrule
\textbf{Method} & \textbf{Acc} & \textbf{F1} & \textbf{Acc} & \textbf{F1} \\
\midrule
DMDP & \textbf{0.810 (0.3)} & \textbf{0.806 (0.4)} & \textbf{0.775 (0.5)} & \textbf{0.774 (0.3)}\\
w UPT & 0.783 (1.6) & 0.771 (1.2) & 0.747 (0.9) & 0.743 (0.8) \\
w IPT & 0.802 (0.6) & 0.799 (0.6) & 0.762 (1.2) & 0.762 (1.2) \\
w/o Gating & 0.803 (0.1) & 0.800 (0.3) & 0.768 (0.3) & 0.767 (0.4) \\
w/o WeightMod & 0.794 (0.7) & 0.798 (0.6) & 0.761 (0.7) & 0.767 (0.6) \\
w V->T PT & 0.799 (1.7) & 0.796 (1.6) & 0.758 (1.7) & 0.752 (1.2) \\
\bottomrule
\end{tabular}
\end{table}
\endgroup

\begingroup
\begin{table}[t]
\caption{\label{tab:cross_dataset_stats} Cross Dataset Statistics}
\centering
\renewcommand{\arraystretch}{1.2} 
\setlength\tabcolsep{10pt} 
\scriptsize 
\begin{tabular}{l c c}
\toprule
\textbf{Dataset} & \textbf{Sarcastic Samples} & \textbf{Non-sarcastic Samples} \\
\midrule
RedEval & 395 & 609 \\ 
MCMD & 183 & 123 \\ 
\bottomrule
\end{tabular}
\end{table}
\endgroup


\subsection{Removing Cross-Modal Prompt Alignment}
In our model DMDP, we project textual prompts to construct visual prompts, enabling effective cross-modal interaction. To demonstrate the effectiveness of this approach, we perform an ablation study by experimenting with a variant of our model that uses independent textual and visual prompts, where the projection layer used to map textual prompt tokens to visual ones is omitted. This variant, referred to as \texttt{w IPT} (row 3), eliminates the coupling between the two modalities during prompt construction. From Table \ref{tab:ablation} ,we observe a drop in performance (Acc: $\downarrow 0.8\%$ in MMSD and $\downarrow 1.3\%$ in MMSD2.0) of this variant. Sarcasm is heavily dependent on the interaction of the two modalities. By projecting textual prompts into the visual space, our approach uses the rich semantic grounding of text to guide and refine the focus of visual prompts. This coupling ensures that the visual prompts are aligned with the textual context, enabling the vision encoder to concentrate on image regions that are most relevant to the underlying semantics of the text. 

\begin{figure}[t]
  
\includegraphics[width=0.43\textwidth, height=0.23\textheight]{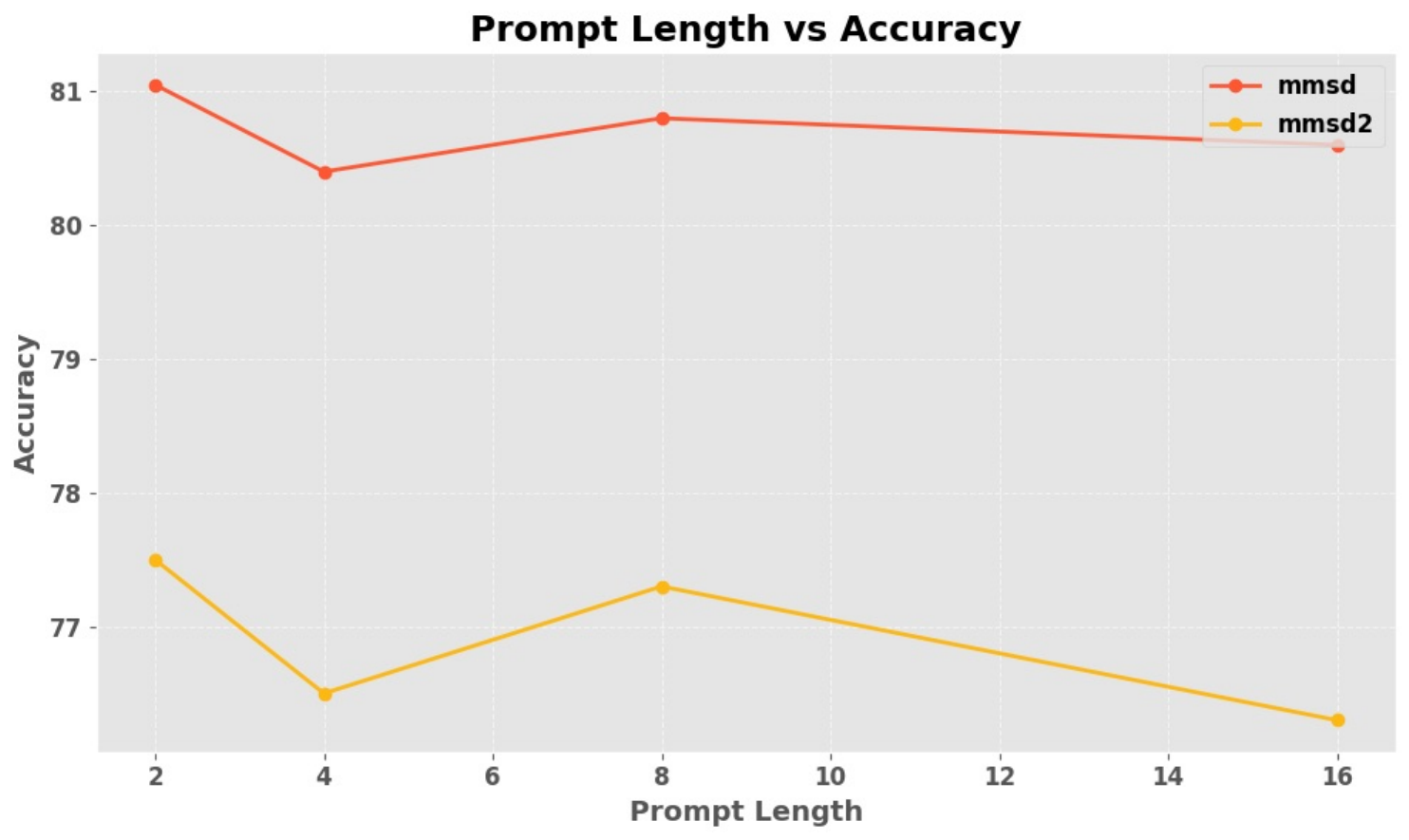}
\caption{Effect of Prompt Length}
\label{fig:length}
\end{figure}

\subsection{Visual to Textual Prompt Tuning}
\label{subsec:reverse}
We ablate DMDP by reversing the construction of prompts, specifically by constructing textual prompts as the projection of visual prompts. From Table \ref{tab:ablation}, we observe a significant drop in performance (Acc: $\downarrow 1.1\%$ in MMSD and $\downarrow 1.7\%$ in MMSD2.0) of this variant \texttt{w V->T PT} (row 6) compared to our original design, where visual prompts are constructed as a projection of textual prompts.

This performance discrepancy highlights the importance of semantic grounding provided by textual information. Text, as a modality, inherently carries rich semantic meaning, often acting as a guide to contextualize and interpret visual information. By projecting textual prompts into the visual space, we enable the model to establish a strong semantic prior, effectively guiding the vision encoder to focus on the most relevant aspects of the image. For instance, if the text describes a specific action or object, the projected prompts can act as a lens for the vision encoder, narrowing its attention to the parts of the image that are critical for understanding the text.

\begingroup
\
\begin{table}[t]
\caption{\label{tab:cross_dataset_perf}
Cross-dataset evaluation. Models trained on 1\% MMSD2.0 dataset and tested on MCMD and RedEval. Numbers in brackets indicate standard deviation.}
\centering
\renewcommand{\arraystretch}{1.3}
\setlength{\tabcolsep}{6pt}
\begin{adjustbox}{width=0.9\linewidth}
\begin{tabular}{l l l l l l}
\toprule
& & \multicolumn{2}{c}{\textbf{MCMD}}
 & \multicolumn{2}{c}{\textbf{RedEval}} \\
\midrule
\textbf{Type} & \multicolumn{1}{l}{\textbf{Method}} & \textbf{Acc} & \textbf{F1} & \textbf{Acc} & \textbf{F1}\\

\midrule
& Att-BERT \cite{pan-etal-2020-modeling} & 0.477 (0.3) & 0.474 (0.1) & 0.461 (1.1) & 0.457 (0.8)\\

 & DIP \cite{Wen2023DIPDI} & 0.545 (1.2) & 0.545 (0.8) & 0.532 (0.2) & 0.513 (0.7) \\

Non-Prompt& DynRT \cite{tian-etal-2023-dynamic} & 0.519 (1.6) & 0.518 (1.4) & 0.541 (0.6) & 0.537 (0.9) \\
Based& MV-CLIP \cite{qin-etal-2023-mmsd2} & 0.653 (0.2) & 0.641 (0.2) & 0.623 (1.1) & 0.620 (1.3) \\
& RAG-LLaVA \cite{Tang2024LeveragingGL} & 0.624 (1.7) & 0.623 (1.5) & 0.617 (1.9) & 0.611 (1.4) \\


\midrule
 & PVLM \cite{Yu2022FewShotMS}  & 0.564 (1.8) & 0.541 (1.3) & 0.553 (1.2) & 0.552 (1.1) \\
Prompt& UP-MPF \cite{Yu2022UnifiedMP} & 0.582 (2.1) & 0.577 (1.9) & 0.569 (0.1) & 0.561 (0.3) \\

Based&CoOp \cite{Zhou2021LearningTP} & 0.663 (0.4) & 0.662 (0.2) & 0.629 (0.9) & 0.618 (0.7) \\

&CoCoOp \cite{Zhou2022ConditionalPL} & 0.658 (1.0) & 0.649 (0.5) & 0.637 (1.2) & 0.631 (1.1)\\

& CAMP & 0.601 (1.3) & 0.591 (1.6) & 0.631 (0.7) & 0.628 (1.2) \\

\midrule

& LLaMA3.2-11B & 0.693 (0.4) & 0.682 (0.3) & 0.641 (0.1) & 0.653 (0.4) \\
LVLM & LLaVA1.6-7B  & 0.674 (0.5) & 0.669 (0.3) & 0.642 (0.4) & 0.641 (0.6) \\
& Qwen2.5-VL-7B  & 0.691 (0.3) & 0.685 (0.6) & 0.656 (0.7) & 0.659 (0.5) \\
\midrule
& DMDP (ours) & \textbf{0.697 (0.3)} & \textbf{0.685 (0.2)} & \textbf{0.663 (0.6)} & \textbf{0.661 (0.5)}\\
\bottomrule
\end{tabular}
\end{adjustbox}
\end{table}
\endgroup

\subsection{Effect of Prompt Length}
Figure \ref{fig:length} shows the effect of varying the number of prompt tokens on the performance of our model. It can be observed that as the prompt length is increased beyond 2, the performance tends to decrease. This suggests that a greater number of prompt tokens may lead to the learning of redundant information, which in turn causes overfitting and degrades the model's performance. As the prompt length increases, the model might focus on irrelevant details or noise, leading to poorer generalization on the task.

\section{Cross-Dataset Evaluation}
To evaluate the generalization ability of DMDP, we conducted experiments on two additional datasets. Since MMSD2.0 \cite{qin-etal-2023-mmsd2} is a more balanced dataset compared to MMSD, we train all multimodal baselines on the 1\% few-shot split of MMSD2.0 and test their performance on these new datasets. One of these datasets is RedEval \cite{Tang2024LeveragingGL} which consists of posts (image + text) collected from Reddit, and the other is a new dataset we introduce, named MCMD (Multi-modal Code-Mixed Memes Dataset), derived from the work of \cite{Maity2022AMF}. Given the limited availability of publicly accessible multimodal sarcasm datasets, we selected this dataset because of its similarity to the task at hand and the presence of labeled sarcasm samples. To construct MCMD, we discard memes lacking sarcasm labels or those containing code-mixed text. The statistics of both of these datasets is reported in Table \ref{tab:cross_dataset_stats}.
From Table \ref{tab:cross_dataset_perf}, it is evident that DMDP demonstrates a stronger generalization ability. 

When compared to non-prompt-based baselines and LVLMs, since DMDP uses the pre-trained knowledge of CLIP using prompts effectively, hence it tends to overfit less on the training data and perform better in cross-dataset transfer. Compared to prompt-based baselines, since we effectively learn the synergy between modalities using deep-modality-disentagled prompts, the image-text relationship is captured effectively, and hence our model DMDP shows superior performance.

\begin{figure}[t]
\centering
\includegraphics[width=0.45\textwidth, height=0.5\textwidth]{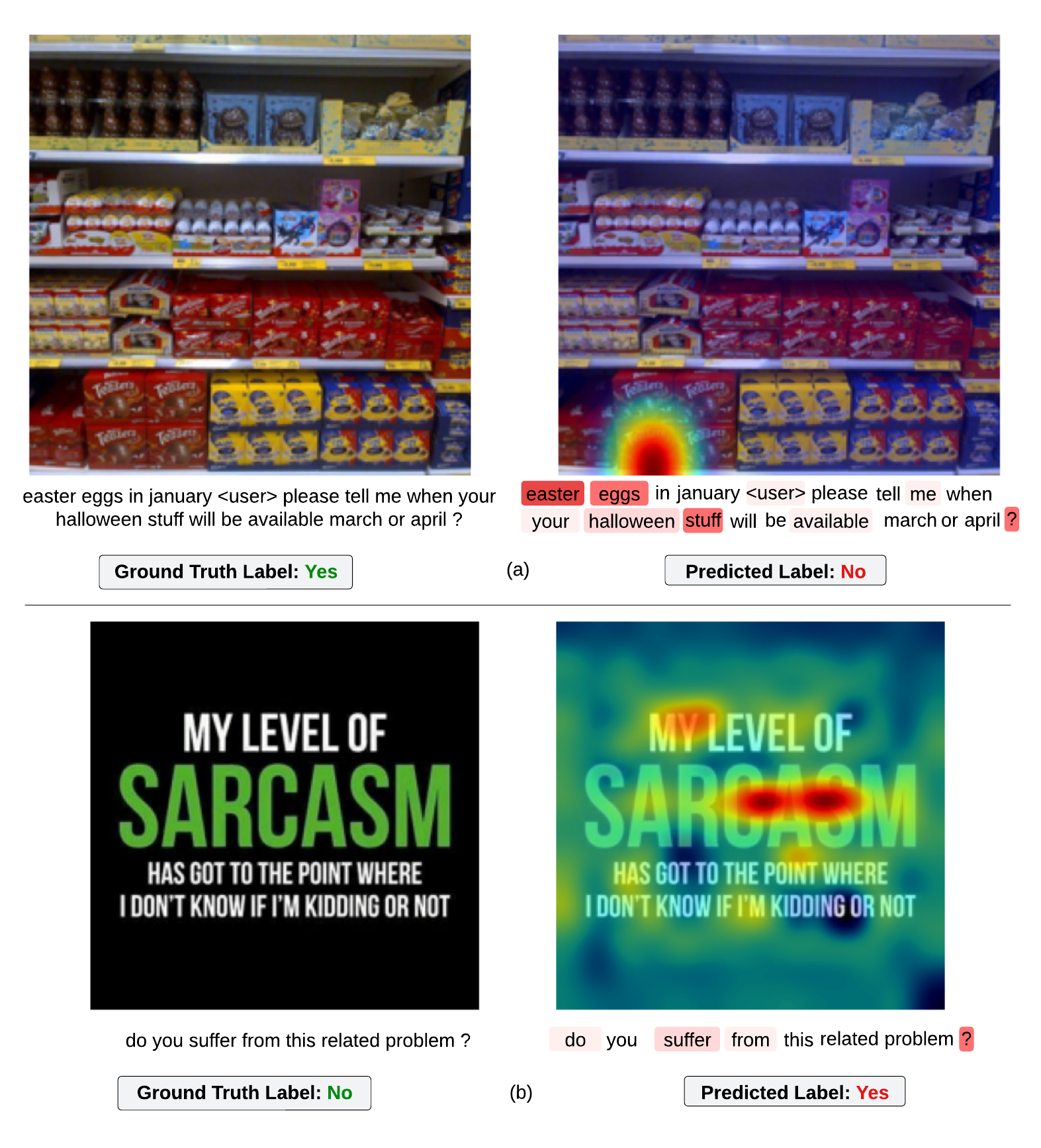} 
\caption{\label{fig:error}Examples of misclassified samples.}

\end{figure}

\section{Error Analysis}
To assess our model's limitations, we conducted error analysis on misclassified samples and identified the following cases:
\begin{itemize}
    \item \textbf{Limited recognition of contextual sarcasm:} As illustrated in Fig.~\ref{fig:error}(a), the sarcastic intent arises from social knowledge that Easter eggs are typically not available in stores during January. The user leverages this context to jokingly ask when Halloween products will be available, implying the absurdity of selling seasonal items out of season. However, the model fails to recognize this contextual cue and instead focuses primarily on literal keywords, such as \textit{easter eggs} in the text and the corresponding products depicted in the image. As a result, it misses the underlying sarcasm that depends on real-world context and the mismatch between expectations and reality.

    \item \textbf{Over-reliance on salient keywords:}
We also observe errors in cases where the model is overly sensitive to certain prominent keywords. As shown in Fig.~\ref{fig:error}(b), the image prominently features the word \textit{sarcasm}. Rather than attempting to interpret the actual meaning in context, the model fixates on this keyword and predicts sarcasm, even though the caption is a genuine question. This indicates that DMDP sometimes gives undue emphasis to superficial textual cues at the expense of deeper semantic understanding of the overall input.
\end{itemize}

\section{Conclusion}
We tackle few-shot multimodal sarcasm detection, where existing baselines struggle due to complex architectures and limited use of pre-trained multimodal knowledge. Existing prompt-learning methods also fall short, as they employ shallow, unified prompts that fail to handle diverse sarcasm types. We introduce DMDP, a deep modality-disentangled prompt tuning approach with cross-modal prompt alignment and layer-wise prompt sharing to better capture text–image interactions. To handle modality dominance, we propose a prompt weighting mechanism, and to stabilize training, we add a gating module that regulates prompt influence early on. Experiments on two public datasets show DMDP outperforms strong baselines in few-shot, extremely low-resource, and OOD scenarios.


\bibliographystyle{ACM-Reference-Format}
\bibliography{sample-base}


\begin{thebibliography}{56}


\ifx \showCODEN    \undefined \def \showCODEN     #1{\unskip}     \fi
\ifx \showISBNx    \undefined \def \showISBNx     #1{\unskip}     \fi
\ifx \showISBNxiii \undefined \def \showISBNxiii  #1{\unskip}     \fi
\ifx \showISSN     \undefined \def \showISSN      #1{\unskip}     \fi
\ifx \showLCCN     \undefined \def \showLCCN      #1{\unskip}     \fi
\ifx \shownote     \undefined \def \shownote      #1{#1}          \fi
\ifx \showarticletitle \undefined \def \showarticletitle #1{#1}   \fi
\ifx \showURL      \undefined \def \showURL       {\relax}        \fi
\providecommand\bibfield[2]{#2}
\providecommand\bibinfo[2]{#2}
\providecommand\natexlab[1]{#1}
\providecommand\showeprint[2][]{arXiv:#2}

\bibitem[Agrawal and An(2018)]%
        {Agrawal2018AffectiveRF}
\bibfield{author}{\bibinfo{person}{Ameeta Agrawal} {and} \bibinfo{person}{Aijun An}.} \bibinfo{year}{2018}\natexlab{}.
\newblock \showarticletitle{Affective Representations for Sarcasm Detection}.
\newblock \bibinfo{journal}{\emph{The 41st International ACM SIGIR Conference on Research \& Development in Information Retrieval}} (\bibinfo{year}{2018}).
\newblock


\bibitem[Agrawal et~al\mbox{.}(2020)]%
        {Agrawal2020LeveragingTO}
\bibfield{author}{\bibinfo{person}{Ameeta Agrawal}, \bibinfo{person}{Aijun An}, {and} \bibinfo{person}{Manos Papagelis}.} \bibinfo{year}{2020}\natexlab{}.
\newblock \showarticletitle{Leveraging Transitions of Emotions for Sarcasm Detection}.
\newblock \bibinfo{journal}{\emph{Proceedings of the 43rd International ACM SIGIR Conference on Research and Development in Information Retrieval}} (\bibinfo{year}{2020}).
\newblock


\bibitem[Amir et~al\mbox{.}(2016)]%
        {amir-etal-2016-modelling}
\bibfield{author}{\bibinfo{person}{Silvio Amir}, \bibinfo{person}{Byron~C. Wallace}, \bibinfo{person}{Hao Lyu}, \bibinfo{person}{Paula Carvalho}, {and} \bibinfo{person}{M{\'a}rio~J. Silva}.} \bibinfo{year}{2016}\natexlab{}.
\newblock \showarticletitle{Modelling Context with User Embeddings for Sarcasm Detection in Social Media}. In \bibinfo{booktitle}{\emph{Proceedings of the 20th {SIGNLL} Conference on Computational Natural Language Learning}}. \bibinfo{address}{Berlin, Germany}.
\newblock


\bibitem[Averbeck(2013)]%
        {Averbeck2013ComparisonsOI}
\bibfield{author}{\bibinfo{person}{Joshua~M. Averbeck}.} \bibinfo{year}{2013}\natexlab{}.
\newblock \showarticletitle{Comparisons of Ironic and Sarcastic Arguments in Terms of Appropriateness and Effectiveness in Personal Relationships}.
\newblock \bibinfo{journal}{\emph{Argumentation and Advocacy}}  \bibinfo{volume}{50} (\bibinfo{year}{2013}), \bibinfo{pages}{47 -- 57}.
\newblock


\bibitem[Babanejad et~al\mbox{.}(2020)]%
        {babanejad-etal-2020-affective}
\bibfield{author}{\bibinfo{person}{Nastaran Babanejad}, \bibinfo{person}{Heidar Davoudi}, \bibinfo{person}{Aijun An}, {and} \bibinfo{person}{Manos Papagelis}.} \bibinfo{year}{2020}\natexlab{}.
\newblock \showarticletitle{Affective and Contextual Embedding for Sarcasm Detection}. In \bibinfo{booktitle}{\emph{Proceedings of the 28th International Conference on Computational Linguistics}}. \bibinfo{address}{Barcelona, Spain (Online)}.
\newblock


\bibitem[Badlani et~al\mbox{.}(2019)]%
        {badlani-etal-2019-ensemble}
\bibfield{author}{\bibinfo{person}{Rohan Badlani}, \bibinfo{person}{Nishit Asnani}, {and} \bibinfo{person}{Manan Rai}.} \bibinfo{year}{2019}\natexlab{}.
\newblock \showarticletitle{An Ensemble of Humour, Sarcasm, and Hate Speechfor Sentiment Classification in Online Reviews}. In \bibinfo{booktitle}{\emph{Proceedings of the 5th Workshop on Noisy User-generated Text (W-NUT 2019)}}, \bibfield{editor}{\bibinfo{person}{Wei Xu}, \bibinfo{person}{Alan Ritter}, \bibinfo{person}{Tim Baldwin}, {and} \bibinfo{person}{Afshin Rahimi}} (Eds.). \bibinfo{publisher}{Association for Computational Linguistics}, \bibinfo{address}{Hong Kong, China}.
\newblock


\bibitem[Cai et~al\mbox{.}(2019)]%
        {cai-etal-2019-multi}
\bibfield{author}{\bibinfo{person}{Yitao Cai}, \bibinfo{person}{Huiyu Cai}, {and} \bibinfo{person}{Xiaojun Wan}.} \bibinfo{year}{2019}\natexlab{}.
\newblock \showarticletitle{Multi-Modal Sarcasm Detection in {T}witter with Hierarchical Fusion Model}. In \bibinfo{booktitle}{\emph{Proceedings of the 57th Annual Meeting of the Association for Computational Linguistics}}, \bibfield{editor}{\bibinfo{person}{Anna Korhonen}, \bibinfo{person}{David Traum}, {and} \bibinfo{person}{Llu{\'\i}s M{\`a}rquez}} (Eds.). \bibinfo{publisher}{Association for Computational Linguistics}, \bibinfo{address}{Florence, Italy}.
\newblock


\bibitem[Cao et~al\mbox{.}(2022)]%
        {cao-etal-2022-prompting}
\bibfield{author}{\bibinfo{person}{Rui Cao}, \bibinfo{person}{Roy Ka-Wei Lee}, \bibinfo{person}{Wen-Haw Chong}, {and} \bibinfo{person}{Jing Jiang}.} \bibinfo{year}{2022}\natexlab{}.
\newblock \showarticletitle{Prompting for Multimodal Hateful Meme Classification}. In \bibinfo{booktitle}{\emph{Proceedings of the 2022 Conference on Empirical Methods in Natural Language Processing}}, \bibfield{editor}{\bibinfo{person}{Yoav Goldberg}, \bibinfo{person}{Zornitsa Kozareva}, {and} \bibinfo{person}{Yue Zhang}} (Eds.). \bibinfo{publisher}{Association for Computational Linguistics}, \bibinfo{address}{Abu Dhabi, United Arab Emirates}.
\newblock


\bibitem[Devlin et~al\mbox{.}(2019)]%
        {Devlin2019BERTPO}
\bibfield{author}{\bibinfo{person}{Jacob Devlin}, \bibinfo{person}{Ming-Wei Chang}, \bibinfo{person}{Kenton Lee}, {and} \bibinfo{person}{Kristina Toutanova}.} \bibinfo{year}{2019}\natexlab{}.
\newblock \showarticletitle{BERT: Pre-training of Deep Bidirectional Transformers for Language Understanding}. In \bibinfo{booktitle}{\emph{North American Chapter of the Association for Computational Linguistics}}.
\newblock


\bibitem[Dosovitskiy et~al\mbox{.}(2020)]%
        {Dosovitskiy2020AnII}
\bibfield{author}{\bibinfo{person}{Alexey Dosovitskiy}, \bibinfo{person}{Lucas Beyer}, \bibinfo{person}{Alexander Kolesnikov}, \bibinfo{person}{Dirk Weissenborn}, \bibinfo{person}{Xiaohua Zhai}, \bibinfo{person}{Thomas Unterthiner}, \bibinfo{person}{Mostafa Dehghani}, \bibinfo{person}{Matthias Minderer}, \bibinfo{person}{Georg Heigold}, \bibinfo{person}{Sylvain Gelly}, \bibinfo{person}{Jakob Uszkoreit}, {and} \bibinfo{person}{Neil Houlsby}.} \bibinfo{year}{2020}\natexlab{}.
\newblock \showarticletitle{An Image is Worth 16x16 Words: Transformers for Image Recognition at Scale}.
\newblock \bibinfo{journal}{\emph{ArXiv}}  \bibinfo{volume}{abs/2010.11929} (\bibinfo{year}{2020}).
\newblock


\bibitem[Gao et~al\mbox{.}(2021)]%
        {gao-etal-2021-making}
\bibfield{author}{\bibinfo{person}{Tianyu Gao}, \bibinfo{person}{Adam Fisch}, {and} \bibinfo{person}{Danqi Chen}.} \bibinfo{year}{2021}\natexlab{}.
\newblock \showarticletitle{Making Pre-trained Language Models Better Few-shot Learners}. In \bibinfo{booktitle}{\emph{Proceedings of the 59th Annual Meeting of the Association for Computational Linguistics and the 11th International Joint Conference on Natural Language Processing (Volume 1: Long Papers)}}, \bibfield{editor}{\bibinfo{person}{Chengqing Zong}, \bibinfo{person}{Fei Xia}, \bibinfo{person}{Wenjie Li}, {and} \bibinfo{person}{Roberto Navigli}} (Eds.). \bibinfo{publisher}{Association for Computational Linguistics}, \bibinfo{address}{Online}.
\newblock


\bibitem[Ghosh et~al\mbox{.}(2017)]%
        {ghosh-etal-2017-role}
\bibfield{author}{\bibinfo{person}{Debanjan Ghosh}, \bibinfo{person}{Alexander Richard~Fabbri}, {and} \bibinfo{person}{Smaranda Muresan}.} \bibinfo{year}{2017}\natexlab{}.
\newblock \showarticletitle{The Role of Conversation Context for Sarcasm Detection in Online Interactions}. In \bibinfo{booktitle}{\emph{Proceedings of the 18th Annual {SIG}dial Meeting on Discourse and Dialogue}}. \bibinfo{address}{Saarbr{\"u}cken, Germany}.
\newblock


\bibitem[Ghosh et~al\mbox{.}(2021)]%
        {ghosh-etal-2021-laughing}
\bibfield{author}{\bibinfo{person}{Debanjan Ghosh}, \bibinfo{person}{Ritvik Shrivastava}, {and} \bibinfo{person}{Smaranda Muresan}.} \bibinfo{year}{2021}\natexlab{}.
\newblock \showarticletitle{{``}Laughing at you or with you{''}: The Role of Sarcasm in Shaping the Disagreement Space}. In \bibinfo{booktitle}{\emph{Proceedings of the 16th Conference of the European Chapter of the Association for Computational Linguistics: Main Volume}}, \bibfield{editor}{\bibinfo{person}{Paola Merlo}, \bibinfo{person}{Jorg Tiedemann}, {and} \bibinfo{person}{Reut Tsarfaty}} (Eds.). \bibinfo{publisher}{Association for Computational Linguistics}, \bibinfo{address}{Online}.
\newblock


\bibitem[Graves and Schmidhuber(2005)]%
        {Graves20052005SI}
\bibfield{author}{\bibinfo{person}{Alex Graves} {and} \bibinfo{person}{J{\"u}rgen Schmidhuber}.} \bibinfo{year}{2005}\natexlab{}.
\newblock \showarticletitle{2005 Special Issue: Framewise phoneme classification with bidirectional LSTM and other neural network architectures}.
\newblock \bibinfo{journal}{\emph{Neural Networks}}  \bibinfo{volume}{18} (\bibinfo{year}{2005}).
\newblock


\bibitem[Han et~al\mbox{.}(2023)]%
        {Han2023E2VPTAE}
\bibfield{author}{\bibinfo{person}{Cheng Han}, \bibinfo{person}{Qifan Wang}, \bibinfo{person}{Yiming Cui}, \bibinfo{person}{Zhiwen Cao}, \bibinfo{person}{Wenguan Wang}, \bibinfo{person}{Siyuan Qi}, {and} \bibinfo{person}{Dongfang Liu}.} \bibinfo{year}{2023}\natexlab{}.
\newblock \showarticletitle{E2VPT: An Effective and Efficient Approach for Visual Prompt Tuning}.
\newblock \bibinfo{journal}{\emph{2023 IEEE/CVF International Conference on Computer Vision (ICCV)}} (\bibinfo{year}{2023}).
\newblock


\bibitem[He et~al\mbox{.}(2015)]%
        {He2015DeepRL}
\bibfield{author}{\bibinfo{person}{Kaiming He}, \bibinfo{person}{X. Zhang}, \bibinfo{person}{Shaoqing Ren}, {and} \bibinfo{person}{Jian Sun}.} \bibinfo{year}{2015}\natexlab{}.
\newblock \showarticletitle{Deep Residual Learning for Image Recognition}.
\newblock \bibinfo{journal}{\emph{2016 IEEE Conference on Computer Vision and Pattern Recognition (CVPR)}} (\bibinfo{year}{2015}).
\newblock


\bibitem[Hegde et~al\mbox{.}(2023)]%
        {Hegde2023CLIPG3}
\bibfield{author}{\bibinfo{person}{Deepti Hegde}, \bibinfo{person}{Jeya Maria~Jose Valanarasu}, {and} \bibinfo{person}{Vishal~M. Patel}.} \bibinfo{year}{2023}\natexlab{}.
\newblock \showarticletitle{CLIP goes 3D: Leveraging Prompt Tuning for Language Grounded 3D Recognition}.
\newblock \bibinfo{journal}{\emph{2023 IEEE/CVF International Conference on Computer Vision Workshops (ICCVW)}} (\bibinfo{year}{2023}).
\newblock


\bibitem[Jana et~al\mbox{.}(2024)]%
        {Jana2024ContinuousAM}
\bibfield{author}{\bibinfo{person}{Soumyadeep Jana}, \bibinfo{person}{Animesh Dey}, {and} \bibinfo{person}{Ranbir Sanasam}.} \bibinfo{year}{2024}\natexlab{}.
\newblock \showarticletitle{Continuous Attentive Multimodal Prompt Tuning for Few-Shot Multimodal Sarcasm Detection}.
\newblock \bibinfo{journal}{\emph{Proceedings of the 28th Conference on Computational Natural Language Learning}} (\bibinfo{year}{2024}).
\newblock


\bibitem[Jia et~al\mbox{.}(2022)]%
        {Jia2022VisualPT}
\bibfield{author}{\bibinfo{person}{Menglin Jia}, \bibinfo{person}{Luming Tang}, \bibinfo{person}{Bor-Chun Chen}, \bibinfo{person}{Claire Cardie}, \bibinfo{person}{Serge~J. Belongie}, \bibinfo{person}{Bharath Hariharan}, {and} \bibinfo{person}{Ser~Nam Lim}.} \bibinfo{year}{2022}\natexlab{}.
\newblock \showarticletitle{Visual Prompt Tuning}.
\newblock \bibinfo{journal}{\emph{ArXiv}} (\bibinfo{year}{2022}).
\newblock


\bibitem[Jian et~al\mbox{.}(2022)]%
        {jian-etal-2022-contrastive}
\bibfield{author}{\bibinfo{person}{Yiren Jian}, \bibinfo{person}{Chongyang Gao}, {and} \bibinfo{person}{Soroush Vosoughi}.} \bibinfo{year}{2022}\natexlab{}.
\newblock \showarticletitle{Contrastive Learning for Prompt-based Few-shot Language Learners}. In \bibinfo{booktitle}{\emph{Proceedings of the 2022 Conference of the North American Chapter of the Association for Computational Linguistics: Human Language Technologies}}, \bibfield{editor}{\bibinfo{person}{Marine Carpuat}, \bibinfo{person}{Marie-Catherine de~Marneffe}, {and} \bibinfo{person}{Ivan~Vladimir Meza~Ruiz}} (Eds.). \bibinfo{publisher}{Association for Computational Linguistics}, \bibinfo{address}{Seattle, United States}.
\newblock


\bibitem[Joshi et~al\mbox{.}(2015)]%
        {Joshi2015HarnessingCI}
\bibfield{author}{\bibinfo{person}{Aditya Joshi}, \bibinfo{person}{Vinita Sharma}, {and} \bibinfo{person}{Pushpak Bhattacharyya}.} \bibinfo{year}{2015}\natexlab{}.
\newblock \showarticletitle{Harnessing Context Incongruity for Sarcasm Detection}. In \bibinfo{booktitle}{\emph{Proceedings of the 53rd Annual Meeting of the Association for Computational Linguistics and the 7th International Joint Conference on Natural Language Processing (Volume 2: Short Papers)}}. \bibinfo{publisher}{Association for Computational Linguistics}, \bibinfo{address}{Beijing, China}.
\newblock


\bibitem[Joshi et~al\mbox{.}(2016)]%
        {joshi-etal-2016-word}
\bibfield{author}{\bibinfo{person}{Aditya Joshi}, \bibinfo{person}{Vaibhav Tripathi}, \bibinfo{person}{Kevin Patel}, \bibinfo{person}{Pushpak Bhattacharyya}, {and} \bibinfo{person}{Mark Carman}.} \bibinfo{year}{2016}\natexlab{}.
\newblock \showarticletitle{Are Word Embedding-based Features Useful for Sarcasm Detection?}. In \bibinfo{booktitle}{\emph{Proceedings of the 2016 Conference on Empirical Methods in Natural Language Processing}}. \bibinfo{address}{Austin, Texas}.
\newblock


\bibitem[Khattri et~al\mbox{.}(2015)]%
        {khattri-etal-2015-sentiment}
\bibfield{author}{\bibinfo{person}{Anupam Khattri}, \bibinfo{person}{Aditya Joshi}, \bibinfo{person}{Pushpak Bhattacharyya}, {and} \bibinfo{person}{Mark Carman}.} \bibinfo{year}{2015}\natexlab{}.
\newblock \showarticletitle{Your Sentiment Precedes You: Using an author{'}s historical tweets to predict sarcasm}. In \bibinfo{booktitle}{\emph{Proceedings of the 6th Workshop on Computational Approaches to Subjectivity, Sentiment and Social Media Analysis}}. \bibinfo{address}{Lisboa, Portugal}.
\newblock


\bibitem[Kim(2014)]%
        {Kim2014ConvolutionalNN}
\bibfield{author}{\bibinfo{person}{Yoon Kim}.} \bibinfo{year}{2014}\natexlab{}.
\newblock \showarticletitle{Convolutional Neural Networks for Sentence Classification}. In \bibinfo{booktitle}{\emph{Conference on Empirical Methods in Natural Language Processing}}.
\newblock


\bibitem[Lester et~al\mbox{.}(2021)]%
        {Lester2021ThePO}
\bibfield{author}{\bibinfo{person}{Brian Lester}, \bibinfo{person}{Rami Al-Rfou}, {and} \bibinfo{person}{Noah Constant}.} \bibinfo{year}{2021}\natexlab{}.
\newblock \showarticletitle{The Power of Scale for Parameter-Efficient Prompt Tuning}. In \bibinfo{booktitle}{\emph{Conference on Empirical Methods in Natural Language Processing}}.
\newblock


\bibitem[Li and Liang(2021)]%
        {Li2021PrefixTuningOC}
\bibfield{author}{\bibinfo{person}{Xiang~Lisa Li} {and} \bibinfo{person}{Percy Liang}.} \bibinfo{year}{2021}\natexlab{}.
\newblock \showarticletitle{Prefix-Tuning: Optimizing Continuous Prompts for Generation}.
\newblock \bibinfo{journal}{\emph{Proceedings of the 59th Annual Meeting of the Association for Computational Linguistics and the 11th International Joint Conference on Natural Language Processing (Volume 1: Long Papers)}}  \bibinfo{volume}{abs/2101.00190} (\bibinfo{year}{2021}).
\newblock


\bibitem[Liang et~al\mbox{.}(2021)]%
        {Liang2021MultiModalSD}
\bibfield{author}{\bibinfo{person}{Bin Liang}, \bibinfo{person}{Chenwei Lou}, \bibinfo{person}{Xiang Li}, \bibinfo{person}{Lin Gui}, \bibinfo{person}{Min Yang}, {and} \bibinfo{person}{Ruifeng Xu}.} \bibinfo{year}{2021}\natexlab{}.
\newblock \showarticletitle{Multi-Modal Sarcasm Detection with Interactive In-Modal and Cross-Modal Graphs}.
\newblock \bibinfo{journal}{\emph{Proceedings of the 29th ACM International Conference on Multimedia}} (\bibinfo{year}{2021}).
\newblock


\bibitem[Liang et~al\mbox{.}(2022)]%
        {Liang2022MultiModalSD}
\bibfield{author}{\bibinfo{person}{Bin Liang}, \bibinfo{person}{Chenwei Lou}, \bibinfo{person}{Xiang Li}, \bibinfo{person}{Min Yang}, \bibinfo{person}{Lin Gui}, \bibinfo{person}{Yulan He}, \bibinfo{person}{Wenjie Pei}, {and} \bibinfo{person}{Ruifeng Xu}.} \bibinfo{year}{2022}\natexlab{}.
\newblock \showarticletitle{Multi-Modal Sarcasm Detection via Cross-Modal Graph Convolutional Network}. In \bibinfo{booktitle}{\emph{Annual Meeting of the Association for Computational Linguistics}}.
\newblock


\bibitem[Liu et~al\mbox{.}(2022a)]%
        {liu-etal-2022-towards-multi-modal}
\bibfield{author}{\bibinfo{person}{Hui Liu}, \bibinfo{person}{Wenya Wang}, {and} \bibinfo{person}{Haoliang Li}.} \bibinfo{year}{2022}\natexlab{a}.
\newblock \showarticletitle{Towards Multi-Modal Sarcasm Detection via Hierarchical Congruity Modeling with Knowledge Enhancement}. In \bibinfo{booktitle}{\emph{Proceedings of the 2022 Conference on Empirical Methods in Natural Language Processing}}, \bibfield{editor}{\bibinfo{person}{Yoav Goldberg}, \bibinfo{person}{Zornitsa Kozareva}, {and} \bibinfo{person}{Yue Zhang}} (Eds.). \bibinfo{publisher}{Association for Computational Linguistics}, \bibinfo{address}{Abu Dhabi, United Arab Emirates}.
\newblock


\bibitem[Liu et~al\mbox{.}(2019)]%
        {Liu2019RoBERTaAR}
\bibfield{author}{\bibinfo{person}{Yinhan Liu}, \bibinfo{person}{Myle Ott}, \bibinfo{person}{Naman Goyal}, \bibinfo{person}{Jingfei Du}, \bibinfo{person}{Mandar Joshi}, \bibinfo{person}{Danqi Chen}, \bibinfo{person}{Omer Levy}, \bibinfo{person}{Mike Lewis}, \bibinfo{person}{Luke Zettlemoyer}, {and} \bibinfo{person}{Veselin Stoyanov}.} \bibinfo{year}{2019}\natexlab{}.
\newblock \showarticletitle{RoBERTa: A Robustly Optimized BERT Pretraining Approach}.
\newblock \bibinfo{journal}{\emph{ArXiv}}  \bibinfo{volume}{abs/1907.11692} (\bibinfo{year}{2019}).
\newblock


\bibitem[Liu et~al\mbox{.}(2022b)]%
        {liu-etal-2022-dual}
\bibfield{author}{\bibinfo{person}{Yiyi Liu}, \bibinfo{person}{Yequan Wang}, \bibinfo{person}{Aixin Sun}, \bibinfo{person}{Xuying Meng}, \bibinfo{person}{Jing Li}, {and} \bibinfo{person}{Jiafeng Guo}.} \bibinfo{year}{2022}\natexlab{b}.
\newblock \showarticletitle{A Dual-Channel Framework for Sarcasm Recognition by Detecting Sentiment Conflict}. In \bibinfo{booktitle}{\emph{Findings of the Association for Computational Linguistics: NAACL 2022}}. \bibinfo{address}{Seattle, United States}.
\newblock


\bibitem[Lou et~al\mbox{.}(2021)]%
        {Lou2021AffectiveDG}
\bibfield{author}{\bibinfo{person}{Chenwei Lou}, \bibinfo{person}{Bin Liang}, \bibinfo{person}{Lin Gui}, \bibinfo{person}{Yulan He}, \bibinfo{person}{Yixue Dang}, {and} \bibinfo{person}{Ruifeng Xu}.} \bibinfo{year}{2021}\natexlab{}.
\newblock \showarticletitle{Affective Dependency Graph for Sarcasm Detection}.
\newblock \bibinfo{journal}{\emph{Proceedings of the 44th International ACM SIGIR Conference on Research and Development in Information Retrieval}} (\bibinfo{year}{2021}).
\newblock


\bibitem[Maity et~al\mbox{.}(2022)]%
        {Maity2022AMF}
\bibfield{author}{\bibinfo{person}{Krishanu Maity}, \bibinfo{person}{Prince Jha}, \bibinfo{person}{Sriparna Saha}, {and} \bibinfo{person}{Pushpak Bhattacharyya}.} \bibinfo{year}{2022}\natexlab{}.
\newblock \showarticletitle{A Multitask Framework for Sentiment, Emotion and Sarcasm aware Cyberbullying Detection from Multi-modal Code-Mixed Memes}.
\newblock \bibinfo{journal}{\emph{Proceedings of the 45th International ACM SIGIR Conference on Research and Development in Information Retrieval}} (\bibinfo{year}{2022}).
\newblock


\bibitem[Maynard and Greenwood(2014)]%
        {maynard-greenwood-2014-cares}
\bibfield{author}{\bibinfo{person}{Diana Maynard} {and} \bibinfo{person}{Mark Greenwood}.} \bibinfo{year}{2014}\natexlab{}.
\newblock \showarticletitle{Who cares about Sarcastic Tweets? Investigating the Impact of Sarcasm on Sentiment Analysis.}. In \bibinfo{booktitle}{\emph{Proceedings of the Ninth International Conference on Language Resources and Evaluation ({LREC}'14)}}. \bibinfo{address}{Reykjavik, Iceland}.
\newblock


\bibitem[Mo et~al\mbox{.}(2024)]%
        {Mo2024LSPTLS}
\bibfield{author}{\bibinfo{person}{Shentong Mo}, \bibinfo{person}{Yansen Wang}, \bibinfo{person}{Xufang Luo}, {and} \bibinfo{person}{Dongsheng Li}.} \bibinfo{year}{2024}\natexlab{}.
\newblock \showarticletitle{LSPT: Long-term Spatial Prompt Tuning for Visual Representation Learning}.
\newblock \bibinfo{journal}{\emph{ArXiv}} (\bibinfo{year}{2024}).
\newblock


\bibitem[Pan et~al\mbox{.}(2020)]%
        {pan-etal-2020-modeling}
\bibfield{author}{\bibinfo{person}{Hongliang Pan}, \bibinfo{person}{Zheng Lin}, \bibinfo{person}{Peng Fu}, \bibinfo{person}{Yatao Qi}, {and} \bibinfo{person}{Weiping Wang}.} \bibinfo{year}{2020}\natexlab{}.
\newblock \showarticletitle{Modeling Intra and Inter-modality Incongruity for Multi-Modal Sarcasm Detection}. In \bibinfo{booktitle}{\emph{Findings of the Association for Computational Linguistics: EMNLP 2020}}, \bibfield{editor}{\bibinfo{person}{Trevor Cohn}, \bibinfo{person}{Yulan He}, {and} \bibinfo{person}{Yang Liu}} (Eds.). \bibinfo{publisher}{Association for Computational Linguistics}, \bibinfo{address}{Online}.
\newblock


\bibitem[Poria et~al\mbox{.}(2016)]%
        {poria-etal-2016-deeper}
\bibfield{author}{\bibinfo{person}{Soujanya Poria}, \bibinfo{person}{Erik Cambria}, \bibinfo{person}{Devamanyu Hazarika}, {and} \bibinfo{person}{Prateek Vij}.} \bibinfo{year}{2016}\natexlab{}.
\newblock \showarticletitle{A Deeper Look into Sarcastic Tweets Using Deep Convolutional Neural Networks}. In \bibinfo{booktitle}{\emph{Proceedings of {COLING} 2016, the 26th International Conference on Computational Linguistics: Technical Papers}}. \bibinfo{address}{Osaka, Japan}.
\newblock


\bibitem[Qin et~al\mbox{.}(2023)]%
        {qin-etal-2023-mmsd2}
\bibfield{author}{\bibinfo{person}{Libo Qin}, \bibinfo{person}{Shijue Huang}, \bibinfo{person}{Qiguang Chen}, \bibinfo{person}{Chenran Cai}, \bibinfo{person}{Yudi Zhang}, \bibinfo{person}{Bin Liang}, \bibinfo{person}{Wanxiang Che}, {and} \bibinfo{person}{Ruifeng Xu}.} \bibinfo{year}{2023}\natexlab{}.
\newblock \showarticletitle{{MMSD}2.0: Towards a Reliable Multi-modal Sarcasm Detection System}. In \bibinfo{booktitle}{\emph{Findings of the Association for Computational Linguistics: ACL 2023}}, \bibfield{editor}{\bibinfo{person}{Anna Rogers}, \bibinfo{person}{Jordan Boyd-Graber}, {and} \bibinfo{person}{Naoaki Okazaki}} (Eds.). \bibinfo{publisher}{Association for Computational Linguistics}, \bibinfo{address}{Toronto, Canada}.
\newblock


\bibitem[Radford et~al\mbox{.}(2021)]%
        {Radford2021LearningTV}
\bibfield{author}{\bibinfo{person}{Alec Radford}, \bibinfo{person}{Jong~Wook Kim}, \bibinfo{person}{Chris Hallacy}, \bibinfo{person}{Aditya Ramesh}, \bibinfo{person}{Gabriel Goh}, \bibinfo{person}{Sandhini Agarwal}, \bibinfo{person}{Girish Sastry}, \bibinfo{person}{Amanda Askell}, \bibinfo{person}{Pamela Mishkin}, \bibinfo{person}{Jack Clark}, \bibinfo{person}{Gretchen Krueger}, {and} \bibinfo{person}{Ilya Sutskever}.} \bibinfo{year}{2021}\natexlab{}.
\newblock \showarticletitle{Learning Transferable Visual Models From Natural Language Supervision}. In \bibinfo{booktitle}{\emph{International Conference on Machine Learning}}.
\newblock


\bibitem[Schifanella et~al\mbox{.}(2016)]%
        {Schifanella2016DetectingSI}
\bibfield{author}{\bibinfo{person}{Rossano Schifanella}, \bibinfo{person}{Paloma de Juan}, \bibinfo{person}{Joel~R. Tetreault}, {and} \bibinfo{person}{Liangliang Cao}.} \bibinfo{year}{2016}\natexlab{}.
\newblock \showarticletitle{Detecting Sarcasm in Multimodal Social Platforms}.
\newblock \bibinfo{journal}{\emph{Proceedings of the 24th ACM international conference on Multimedia}} (\bibinfo{year}{2016}).
\newblock


\bibitem[Tang et~al\mbox{.}(2024)]%
        {Tang2024LeveragingGL}
\bibfield{author}{\bibinfo{person}{Binghao Tang}, \bibinfo{person}{Boda Lin}, \bibinfo{person}{Haolong Yan}, {and} \bibinfo{person}{Si Li}.} \bibinfo{year}{2024}\natexlab{}.
\newblock \showarticletitle{Leveraging Generative Large Language Models with Visual Instruction and Demonstration Retrieval for Multimodal Sarcasm Detection}. In \bibinfo{booktitle}{\emph{North American Chapter of the Association for Computational Linguistics}}.
\newblock


\bibitem[Tian et~al\mbox{.}(2023)]%
        {tian-etal-2023-dynamic}
\bibfield{author}{\bibinfo{person}{Yuan Tian}, \bibinfo{person}{Nan Xu}, \bibinfo{person}{Ruike Zhang}, {and} \bibinfo{person}{Wenji Mao}.} \bibinfo{year}{2023}\natexlab{}.
\newblock \showarticletitle{Dynamic Routing Transformer Network for Multimodal Sarcasm Detection}. In \bibinfo{booktitle}{\emph{Proceedings of the 61st Annual Meeting of the Association for Computational Linguistics (Volume 1: Long Papers)}}, \bibfield{editor}{\bibinfo{person}{Anna Rogers}, \bibinfo{person}{Jordan Boyd-Graber}, {and} \bibinfo{person}{Naoaki Okazaki}} (Eds.). \bibinfo{publisher}{Association for Computational Linguistics}, \bibinfo{address}{Toronto, Canada}.
\newblock


\bibitem[Tindale and Gough(1987)]%
        {Tindale1987TheUO}
\bibfield{author}{\bibinfo{person}{Christopher~W. Tindale} {and} \bibinfo{person}{James Gough}.} \bibinfo{year}{1987}\natexlab{}.
\newblock \showarticletitle{The Use of Irony in Argumentation.}
\newblock \bibinfo{journal}{\emph{Philosophy and Rhetoric}}  \bibinfo{volume}{20} (\bibinfo{year}{1987}), \bibinfo{pages}{1--17}.
\newblock


\bibitem[Wang et~al\mbox{.}(2024)]%
        {Wang2024ViLTCLIPVA}
\bibfield{author}{\bibinfo{person}{Hao Wang}, \bibinfo{person}{Fang Liu}, \bibinfo{person}{Licheng Jiao}, \bibinfo{person}{Jiahao Wang}, \bibinfo{person}{Zehua Hao}, \bibinfo{person}{Shuo Li}, \bibinfo{person}{Lingling Li}, \bibinfo{person}{Puhua Chen}, {and} \bibinfo{person}{Xu Liu}.} \bibinfo{year}{2024}\natexlab{}.
\newblock \showarticletitle{ViLT-CLIP: Video and Language Tuning CLIP with Multimodal Prompt Learning and Scenario-Guided Optimization}. In \bibinfo{booktitle}{\emph{AAAI Conference on Artificial Intelligence}}.
\newblock


\bibitem[Wang et~al\mbox{.}(2023)]%
        {Wang2023SeeingIF}
\bibfield{author}{\bibinfo{person}{Qianqian Wang}, \bibinfo{person}{Junlong Du}, \bibinfo{person}{Ke Yan}, {and} \bibinfo{person}{Shouhong Ding}.} \bibinfo{year}{2023}\natexlab{}.
\newblock \showarticletitle{Seeing in Flowing: Adapting CLIP for Action Recognition with Motion Prompts Learning}.
\newblock \bibinfo{journal}{\emph{Proceedings of the 31st ACM International Conference on Multimedia}} (\bibinfo{year}{2023}).
\newblock


\bibitem[Wasim et~al\mbox{.}(2023)]%
        {Wasim2023VitaCLIPVA}
\bibfield{author}{\bibinfo{person}{Syed~Talal Wasim}, \bibinfo{person}{Muzammal Naseer}, \bibinfo{person}{Salman~H. Khan}, \bibinfo{person}{Fahad~Shahbaz Khan}, {and} \bibinfo{person}{Mubarak Shah}.} \bibinfo{year}{2023}\natexlab{}.
\newblock \showarticletitle{Vita-CLIP: Video and text adaptive CLIP via Multimodal Prompting}.
\newblock \bibinfo{journal}{\emph{2023 IEEE/CVF Conference on Computer Vision and Pattern Recognition (CVPR)}} (\bibinfo{year}{2023}).
\newblock


\bibitem[Wen et~al\mbox{.}(2023)]%
        {Wen2023DIPDI}
\bibfield{author}{\bibinfo{person}{Chan~Shao Wen}, \bibinfo{person}{Guoli Jia}, {and} \bibinfo{person}{Jufeng Yang}.} \bibinfo{year}{2023}\natexlab{}.
\newblock \showarticletitle{DIP: Dual Incongruity Perceiving Network for Sarcasm Detection}.
\newblock \bibinfo{journal}{\emph{2023 IEEE/CVF Conference on Computer Vision and Pattern Recognition (CVPR)}} (\bibinfo{year}{2023}).
\newblock


\bibitem[Xing et~al\mbox{.}(2023)]%
        {Xing2023MultimodalAO}
\bibfield{author}{\bibinfo{person}{Jiazheng Xing}, \bibinfo{person}{Mengmeng Wang}, \bibinfo{person}{Xiaojun Hou}, \bibinfo{person}{Guangwen Dai}, \bibinfo{person}{Jingdong Wang}, {and} \bibinfo{person}{Yong Liu}.} \bibinfo{year}{2023}\natexlab{}.
\newblock \showarticletitle{Multimodal Adaptation of CLIP for Few-Shot Action Recognition}.
\newblock \bibinfo{journal}{\emph{ArXiv}} (\bibinfo{year}{2023}).
\newblock


\bibitem[Xu et~al\mbox{.}(2020)]%
        {xu-etal-2020-reasoning}
\bibfield{author}{\bibinfo{person}{Nan Xu}, \bibinfo{person}{Zhixiong Zeng}, {and} \bibinfo{person}{Wenji Mao}.} \bibinfo{year}{2020}\natexlab{}.
\newblock \showarticletitle{Reasoning with Multimodal Sarcastic Tweets via Modeling Cross-Modality Contrast and Semantic Association}. In \bibinfo{booktitle}{\emph{Proceedings of the 58th Annual Meeting of the Association for Computational Linguistics}}, \bibfield{editor}{\bibinfo{person}{Dan Jurafsky}, \bibinfo{person}{Joyce Chai}, \bibinfo{person}{Natalie Schluter}, {and} \bibinfo{person}{Joel Tetreault}} (Eds.). \bibinfo{publisher}{Association for Computational Linguistics}, \bibinfo{address}{Online}.
\newblock


\bibitem[Yang et~al\mbox{.}(2022)]%
        {Yang2022FewshotMS}
\bibfield{author}{\bibinfo{person}{Xiaocui Yang}, \bibinfo{person}{Shi Feng}, \bibinfo{person}{Daling Wang}, \bibinfo{person}{Pengfei Hong}, {and} \bibinfo{person}{Soujanya Poria}.} \bibinfo{year}{2022}\natexlab{}.
\newblock \showarticletitle{Few-shot Multimodal Sentiment Analysis Based on Multimodal Probabilistic Fusion Prompts}.
\newblock \bibinfo{journal}{\emph{Proceedings of the 31st ACM International Conference on Multimedia}} (\bibinfo{year}{2022}).
\newblock


\bibitem[Yoo et~al\mbox{.}(2023)]%
        {Yoo2023ImprovingVP}
\bibfield{author}{\bibinfo{person}{Seung~Woo Yoo}, \bibinfo{person}{Eunji Kim}, \bibinfo{person}{Dahuin Jung}, \bibinfo{person}{Jungbeom Lee}, {and} \bibinfo{person}{Sung-Hoon Yoon}.} \bibinfo{year}{2023}\natexlab{}.
\newblock \showarticletitle{Improving Visual Prompt Tuning for Self-supervised Vision Transformers}.
\newblock \bibinfo{journal}{\emph{ArXiv}} (\bibinfo{year}{2023}).
\newblock


\bibitem[Yu and Zhang(2022)]%
        {Yu2022FewShotMS}
\bibfield{author}{\bibinfo{person}{Yang Yu} {and} \bibinfo{person}{Dong Zhang}.} \bibinfo{year}{2022}\natexlab{}.
\newblock \showarticletitle{Few-Shot Multi-Modal Sentiment Analysis with Prompt-Based Vision-Aware Language Modeling}.
\newblock \bibinfo{journal}{\emph{2022 IEEE International Conference on Multimedia and Expo (ICME)}} (\bibinfo{year}{2022}).
\newblock


\bibitem[Yu et~al\mbox{.}(2022)]%
        {Yu2022UnifiedMP}
\bibfield{author}{\bibinfo{person}{Yang Yu}, \bibinfo{person}{Dong Zhang}, {and} \bibinfo{person}{Shoushan Li}.} \bibinfo{year}{2022}\natexlab{}.
\newblock \showarticletitle{Unified Multi-modal Pre-training for Few-shot Sentiment Analysis with Prompt-based Learning}.
\newblock \bibinfo{journal}{\emph{Proceedings of the 30th ACM International Conference on Multimedia}} (\bibinfo{year}{2022}).
\newblock


\bibitem[Zhang et~al\mbox{.}(2016)]%
        {zhang-etal-2016-tweet}
\bibfield{author}{\bibinfo{person}{Meishan Zhang}, \bibinfo{person}{Yue Zhang}, {and} \bibinfo{person}{Guohong Fu}.} \bibinfo{year}{2016}\natexlab{}.
\newblock \showarticletitle{Tweet Sarcasm Detection Using Deep Neural Network}. In \bibinfo{booktitle}{\emph{Proceedings of {COLING} 2016, the 26th International Conference on Computational Linguistics: Technical Papers}}. \bibinfo{address}{Osaka, Japan}.
\newblock


\bibitem[Zhou et~al\mbox{.}(2021)]%
        {Zhou2021LearningTP}
\bibfield{author}{\bibinfo{person}{Kaiyang Zhou}, \bibinfo{person}{Jingkang Yang}, \bibinfo{person}{Chen~Change Loy}, {and} \bibinfo{person}{Ziwei Liu}.} \bibinfo{year}{2021}\natexlab{}.
\newblock \showarticletitle{Learning to Prompt for Vision-Language Models}.
\newblock \bibinfo{journal}{\emph{International Journal of Computer Vision}} (\bibinfo{year}{2021}).
\newblock


\bibitem[Zhou et~al\mbox{.}(2022)]%
        {Zhou2022ConditionalPL}
\bibfield{author}{\bibinfo{person}{Kaiyang Zhou}, \bibinfo{person}{Jingkang Yang}, \bibinfo{person}{Chen~Change Loy}, {and} \bibinfo{person}{Ziwei Liu}.} \bibinfo{year}{2022}\natexlab{}.
\newblock \showarticletitle{Conditional Prompt Learning for Vision-Language Models}.
\newblock \bibinfo{journal}{\emph{2022 IEEE/CVF Conference on Computer Vision and Pattern Recognition (CVPR)}} (\bibinfo{year}{2022}).
\newblock


\end{thebibliography}

\appendix


\end{document}